\documentclass[acmsmall,screen,nonacm]{acmart}

\usepackage{utils}

\AtBeginDocument{%
  }

\setcopyright{none}
\copyrightyear{2025}
\acmYear{2025}
\acmDOI{}
\acmISBN{}

\acmVolume{}
\acmNumber{}
\acmArticle{}
\acmMonth{}

\begin{document}



\title{CatCMA with Margin for Single- and Multi-Objective Mixed-Variable Black-Box Optimization}

\author{Ryoki Hamano}
\email{hamano_ryoki_xa@cyberagent.co.jp}
\orcid{0000-0002-4425-1683}
\affiliation{%
  \institution{CyberAgent}
  \city{Shibuya}
  \state{Tokyo}
  \country{Japan}
  \postcode{150-0042}
}

\author{Masahiro Nomura}
\email{nomura@comp.isct.ac.jp}
\orcid{0000-0002-4945-5984}
\affiliation{%
  \institution{Institute of Science Tokyo}
  \city{Yokohama}
  \state{Kanagawa}
  \country{Japan}
  \postcode{226-0026}
}

\author{Shota Saito}
\email{saito-shota-bt@ynu.jp}
\orcid{0000-0002-9863-6765}
\affiliation{%
  \institution{Yokohama National University}
  \city{Yokohama}
  \state{Kanagawa}
  \country{Japan}
  \postcode{240-8501}
}
\affiliation{%
  \institution{SKILLUP NeXt Ltd.}
  \city{Chiyoda}
  \state{Tokyo}
  \country{Japan}
  \postcode{101-0051}
}

\author{Kento Uchida}
\email{uchida-kento-fz@ynu.ac.jp}
\orcid{0000-0002-4179-6020}
\affiliation{%
  \institution{Yokohama National University}
  \city{Yokohama}
  \state{Kanagawa}
  \country{Japan}
  \postcode{240-8501}
}

\author{Shinichi Shirakawa}
\email{shirakawa-shinichi-bg@ynu.ac.jp}
\orcid{0000-0002-4659-6108}
\affiliation{%
  \institution{Yokohama National University}
  \city{Yokohama}
  \state{Kanagawa}
  \country{Japan}
  \postcode{240-8501}
}

\renewcommand{\shortauthors}{Hamano et al.}

\begin{abstract}


This study focuses on mixed-variable black-box optimization (MV-BBO), addressing continuous, integer, and categorical variables.
Many real-world MV-BBO problems involve dependencies among these different types of variables, requiring efficient methods to optimize them simultaneously.
Recently, stochastic optimization methods leveraging the mechanism of the covariance matrix adaptation evolution strategy have shown promising results in mixed-integer or mixed-category optimization.
However, such methods cannot handle the three types of variables simultaneously.
In this study, we propose CatCMA with Margin (CatCMAwM), a stochastic optimization method for MV-BBO that jointly optimizes continuous, integer, and categorical variables.
CatCMAwM is developed by incorporating novel integer handling into CatCMA, a mixed-category black-box optimization method employing a joint distribution of multivariate Gaussian and categorical distributions.
The proposed integer handling is carefully designed by reviewing existing integer handling and following the design principles of CatCMA.
Furthermore, we extend CatCMAwM to multi-objective MV-BBO by instantiating it within the Sofomore framework, obtaining a multi-objective optimizer termed COMO-CatCMA with Margin (COMO-CatCMAwM).
Numerical experiments on single-objective MV-BBO problems show that CatCMAwM effectively handles the three types of variables, outperforming state-of-the-art Bayesian optimization methods and baselines that simply incorporate existing integer handlings into CatCMA.
Moreover, on bi-objective MV-BBO benchmarks, COMO-CatCMAwM achieves competitive or superior hypervolume compared to representative baselines.

\end{abstract}




\maketitle

\section{Introduction}
Mixed-variable black-box optimization (MV-BBO) involves the simultaneous optimization of different types of variables, such as continuous, integer, and categorical variables, in black-box settings.
MV-BBO problems often appear in various real-world applications, such as hyperparameter optimization in machine learning~\cite{hazan2018hyperparameter, hyperparam_tune:hutter2019}, hardware design~\cite{hardware:lu2020, hardware:touloupas2022}, and development of new materials~\cite{iyer_data_2020, zhang_bayesian_2020}.
In these problems, there are often dependencies among different types of variables, requiring efficient methods to optimize them simultaneously.
However, most black-box optimization methods have been developed for variables of the same type, making it difficult to efficiently combine them for MV-BBO problems.

Previously, natural solutions to MV-BBO problems have been limited to approaches such as those based on genetic algorithms~\cite{GeneAS:1996}, ant colony optimization~\cite{ACO:2014}, and Bayesian optimization~\cite{TPE:2011, MV-BO:2021, SMAC:2011, casmopolitan:2021, SMAC3:2022}.
Recently, mixed-integer black-box optimization (MI-BBO) methods have been successfully developed by incorporating integer handling into existing continuous black-box optimization methods~\cite{hansen_cma-es_2011, CMAwM:2022, 1+1-CMA-ESwM:2023, MI-NES:2023, CMAwM-TELO:2024, LB+IC-CMA-ES:2024}.
Additionally, the mixed-category black-box optimization (MC-BBO) method, referred to as CatCMA~\cite{CatCMA:2024}, has been proposed by utilizing the joint distribution of multivariate Gaussian and categorical distributions as a search distribution. 
Most of these approaches achieve effective optimization for continuous variables by leveraging the mechanism of the covariance matrix adaptation evolution strategy~(CMA-ES)~\cite{hansen_adapting_1996, hansen_reducing_2003,hansen2016cma}.
However, such approaches utilizing the CMA-ES mechanism cannot naturally handle the three types of variables simultaneously.

In this study, we propose CatCMA with Margin (CatCMAwM), a stochastic optimization method for MV-BBO that jointly optimizes continuous, integer, and categorical variables.
CatCMAwM is developed by incorporating a novel integer handling, a modification of the margin correction in CMA-ES with Margin~\cite{CMAwM:2022}, into the multivariate Gaussian distribution of CatCMA.
While the integration of integer handling into CatCMA is straightforward, the challenge lies in establishing a configuration that allows for the efficient optimization of various MV-BBO problems.
To achieve this goal, we first review existing integer handling methods and discuss the requirements for their integration into CatCMA.
Then, we propose a novel integer handling carefully designed following the design principles of CatCMA.
Whereas existing integer handling methods impose only a lower bound on the marginal probability of integer variables, the proposed method imposes both a lower bound and, when necessary, an upper bound.
This upper bound allows us to analytically derive promising settings in CatCMAwM and also enhances the convergence performance of continuous variables.
We evaluate the effectiveness of the proposed integer handling on mixed integer benchmark functions.
Finally, we compare the performance of CatCMAwM on MV-BBO benchmark functions with state-of-the-art Bayesian optimization methods and baselines that simply incorporate existing integer handling into CatCMA.

This manuscript is an extension of the previous work~\cite{catcmawm:2025}, which proposes CatCMAwM for single-objective MV-BBO.
In addition to the single-objective setting, we address multi-objective MV-BBO (MO-MV-BBO).
We extend CatCMAwM to the multi-objective setting by instantiating it within the Sofomore framework~\cite{COMO-CMA-ES:2019}, a general mechanism for constructing multi-objective optimizers from multiple single-objective algorithms without modifying their internal search operators.
We evaluate the resulting multi-objective optimizer, COMO-CatCMA with Margin (COMO-CatCMAwM), on bi-objective mixed-variable benchmarks and show that it attains competitive or superior hypervolume and computational efficiency compared to representative baselines.

%

\section{Preliminaries for Single-Objective MV-BBO}

\subsection{Mixed-Variable Black-Box Optimization} \label{ssec:MV-BBO}
We consider the mixed-variable function $f$ which has $\Nco$ continuous variables, $\Nin$ integer variables, and $\Nca$ categorical variables. 
The search space of $f$ is given by $\mathcal{X} \times \mathcal{Z} \times \mathcal{C}$, where $\mathcal{X} = \R^{\Nco}$, $\mathcal{Z} = \mathcal{Z}_1 \times \cdots \times \mathcal{Z}_{\Nin}$, and $\mathcal{C} = \mathcal{C}_1 \times \cdots \times \mathcal{C}_{\Nca}$.
The $n$-th integer domain $\mathcal{Z}_n$ and the $n$-th categorical domain $\mathcal{C}_n$ are defined as
\begin{align}
    \mathcal{Z}_n &= \left\{ z_{n,1}, \ldots, z_{n,L_n} \right\} \enspace, \\
    \mathcal{C}_n &= \left\{ \bc_n \in \{0, 1\}^{K_n} \enspace \middle| \enspace \sum_{k=1}^{K_n} \bc_{n, k} = 1 \right\} \enspace,
\end{align}
where $L_n$ is the number of elements in the $n$-th integer variable and $K_n$ is the number of categories in the $n$-th categorical variable.
We note that $z_{n,l}$ is the $l$-th smallest integer in $\mathcal{Z}_n$, and $\bc_n$ is a $K_n$-dimensional one-hot vector.
In this study, the integer domain $\mathcal{Z}_n$ can include discrete non-integer values, as in $\mathcal{Z}_n = \{0.01, 0.1, 1\}$.

\subsection{CMA-ES with Margin} \label{ssec:cmawm}
The CMA-ES with Margin~(CMA-ESwM)~\cite{CMAwM:2022} is one of the promising methods for mixed-integer black-box optimization based on the CMA-ES.
The CMA-ESwM employs a multivariate Gaussian distribution $\mathcal{N}(\m[t], (\sig[t])^2 \A[t] \C[t] \A[t])$, where $\m[t]  \in \R^{\Nmi := \Nco + \Nin}$ is the mean vector, $\sig[t] \in \R_{>0}$ is the step-size, $\A[t] \in \R^{\Nmi \times \Nmi}$ is a diagonal matrix with positive diagonal elements, and $\C[t] \in \R^{\Nmi \times \Nmi}$ is the covariance matrix.
The CMA-ESwM repeats the update of the distribution parameters based on the update rules of the CMA-ES and the margin correction, which is an integer handling for the multivariate Gaussian distribution.

\subsubsection{Update Rules Based on the CMA-ES} \label{sssec:update_based_on_cmaes}
~ The CMA-ESwM generates $\lambda$ candidate solutions $(\x_1, \z_1), \ldots, (\x_\lambda, \z_\lambda)$, pairs of continuous and integer samples, at each iteration according to the multivariate Gaussian distribution.
First, we transform samples $\bxi_1, \ldots, \bxi_\lambda$ generated from the $\Nmi$-dimensional standard normal distribution $\mathcal{N}(\boldsymbol{0}, \mathbf{I}_\Nmi)$.
\begin{align}
    \y_i &= \sqrt{\C[t]} \bxi_i \\
    \bv_i &= \m[t] + \sig[t] \A[t] \y_i
\end{align}
The candidate solutions are obtained by discretizing some dimensions of $\bv_1, \ldots, \bv_\lambda$.
In the index set $\{1, \ldots, \Nmi \}$, let $\jco_n$ be the index of the $n$-th continuous variable and let $\jin_n$ be the index of the $n$-th integer variable. Then, we note that $\{ \jco_1, \ldots, \jco_{\Nco} \} \cup \{ \jin_1, \ldots, \jin_{\Nin} \} = \{ 1, \ldots, \Nmi \}$.
The discretization is performed by the encoding function $\Enc$ defined as follows:
\begin{align*}
    [\Enc(\bv)]_{\jco_n} &= [\bv]_{\jco_n} \quad \text{for} \enspace n \in \{1, \ldots, \Nco \} \enspace, \\
    [\Enc(\bv)]_{\jin_n} &= \begin{cases}
        z_{n,1}  &\text{if} \enspace [\bv]_{\jin_n} \leq \ell_{n, 1|2}\\
        z_{n,l}  &\text{if} \enspace \ell_{n, l-1|l} < [\bv]_{\jin_n} \leq \ell_{n, l|l+1}\\
        z_{n,L_n}  &\text{if} \enspace \ell_{n, L_n-1|L_n} < [\bv]_{\jin_n}
    \end{cases}
    \quad \text{for} \enspace n \in \{1, \ldots, \Nin \}
\end{align*}
We denote the $n$-th element of a vector $\bv$ as $[\bv]_n$ and define the threshold between the two consecutive integers $z_{n, l}$ and $z_{n, l+1}$ as $\ell_{n, l|l+1} = (z_{n, l} + z_{n, l+1})/2$.
Then, each element of the candidate solutions is obtained as follows:
\begin{align}
    [\x_i]_n &= [\Enc(\bv_i)]_{\jco_n} \quad \text{for} \enspace n \in \{1, \ldots, \Nco \} \\
    [\z_i]_n &= [\Enc(\bv_i)]_{\jin_n} \quad \text{for} \enspace n \in \{1, \ldots, \Nin \}
\end{align}
The candidate solutions $(\x_1, \z_1), \ldots, (\x_\lambda, \z_\lambda)$ are evaluated by the objective function $f$ and ranked according to their evaluations.

The mean vector is updated as
\begin{align}
    \m[t+1] = \m[t] + c_m \sig[t] \A[t] \sum_{i=1}^\mu w_i \y_{i:\lambda} \enspace, \label{eq:update_m}
\end{align}
where $c_m$ is the learning rate, $i \! :\! \lambda$ denotes the index of the $i$-th best sample, and the weight $w_i$ satisfies $w_1  \geq \cdots \geq  w_\mu  >  0$ and $\sum_{i=1}^\mu  w_i  =  1$.

The evolution paths are updated as
\begin{align}
    \ps[t+1] &= (1-c_\sigma) \ps[t] + \sqrt{c_\sigma(2-c_\sigma)\muw} (\C[t])^{ -\frac12}  \sum_{i=1}^\mu w_i \y_{i:\lambda} ~, \label{eq:update_ps} \\
    \pc[t+1] &= (1-c_c) \pc[t] + h_\sigma \sqrt{c_c(2-c_c)\muw} \sum_{i=1}^\mu w_i \y_{i:\lambda} ~, \label{eq:update_pc}
\end{align}
where $c_\sigma$ and $c_c$ are cumulative rates and $\muw$ denotes $(\sum_{i=1}^\mu w^2_i)^{-1}$. The Heaviside function $h_\sigma$ takes $1$ if it holds
\begin{align}
    \frac{\| \ps[t+1] \|^2}{\sqrt{1 - (1 - c_\sigma)^{2(t+1)}}} < \left( 1.4 + \frac{2}{\Nmi + 1} \right) \E \left[ \left\| \mathcal{N} \left(\boldsymbol{0}, \mathbf{I}_{\Nmi} \right) \right\| \right] \enspace,
\end{align}
and $h_\sigma$ takes $0$ otherwise.
We note that the expected norm $\E \left[ \left\| \mathcal{N} \left(\boldsymbol{0}, \mathbf{I}_{\Nmi} \right) \right\| \right]$ is approximated by $\sqrt{\Nmi}\left( 1 - \frac{1}{4\Nmi} + \frac{1}{21\Nmi^2} \right)$.

The covariance matrix is updated as
\begin{align}
    \begin{split}
    \C[t+1] &= \left( 1 - c_1 - c_\mu \sum_{i=1}^\lambda w_i \bigl. + (1-h_\sigma) c_1 c_c (2-c_c) \right) \C[t] \\
    &\qquad\qquad\qquad\qquad + c_1 \pc[t+1](\pc[t+1])^\top+ c_\mu \sum_{i=1}^\lambda w^\circ_i \y_{i:\lambda} \y_{i:\lambda}^\top \enspace,
    \end{split} \label{eq:update_C}
\end{align}
where $c_1$ and $c_\mu$ are learning rates and $w^\circ_i$ is given as follows:
\begin{align}
    w^\circ_i = 
    \begin{cases}
        w_i \quad &\text{if} \enspace w_i \geq 0 \\
        w_i \cdot \frac{\Nmi}{\left\| (\C[t])^{-\frac12} \y_{i:\lambda} \right\|^2} \quad &\text{otherwise}
    \end{cases} \label{eq:negative_weight}
\end{align}

The step-size is updated with cumulative step-size adaptation as
\begin{align}
    \sig[t+1] = \sig[t] \exp \left( \frac{c_\sigma}{d_\sigma}  \left( \frac{\| \ps[t+1] \|}{\E [\| \mathcal{N} (\boldsymbol{0}, \mathbf{I}_{\Nmi}) \|]} - 1 \right) \right) \enspace, \label{eq:update_sig}
\end{align}
where $d_\sigma$ is the damping factor.

\subsubsection{Margin Correction} \label{sssec:margin_correction}
~ In a certain integer dimension, when the standard deviation of the multivariate Gaussian distribution becomes small compared to the interval of integers, the generated integer variable is fixed to a single value.
To address this problem, the CMA-ESwM introduces a lower bound on the marginal probabilities associated with the generation of integer variables in the multivariate Gaussian distribution, called margin correction.
For each $n \in \{1, \ldots, \Nin\}$, the margin correction updates $[\m[t+1]]_{\jin_n}$ and $\langle \A[t] \rangle_{\jin_n}$, where $\langle \cdot \rangle_{n}$ denotes the $n$-th diagonal element of a matrix.

\paragraph{Case of $~[\textnormal{\textsc{Enc}}(\m[t+1])]_{\jin_n} \in \{ z_{n,1}, z_{n,L_n} \}$.}
Let $\lnei{t+1}{n}$ be the nearest threshold to $\jin_n$-th element of the mean vector. First, the marginal probability $\pmut{t+1}{n}$ is calculated as follows:
\begin{align}
    \pmut{t+1}{n} = &\min\left\{ \Pr\left( [\bv]_{\jin_n} \leq \lnei{t+1}{n} \right),  \Pr\left( \lnei{t+1}{n} < [\bv]_{\jin_n} \right) \right\}
\end{align}
Next, we restrict $\pmut{t+1}{n}$ as
\begin{align}
    \pmut{t+1}{n} \leftarrow \max \left\{ \alpha, \pmut{t+1}{n} \right\} \enspace, \label{eq:clip_edgecase}
\end{align}
where $\alpha$ is the margin value that controls the strength of the correction.
Finally, we modify the mean vector so that the marginal probability becomes restricted $\pmut{t+1}{n}$ as
\begin{align}
    \begin{split}
        &[\m[t+1]]_{\jin_n} \leftarrow \rev{\lnei{t+1}{n}} + \sign \left( [\m[t+1]]_{\jin_n} - \rev{\lnei{t+1}{n}} \right) \\
        &\qquad\qquad\qquad\qquad\qquad \cdot \sig[t+1] \langle \A[t] \rangle_{\jin_n} \sqrt{\langle \C[t+1] \rangle_{\jin_n}\Xppf{1-2\pmut{t+1}{n}}} \enspace,
    \end{split}
\end{align}
where $\Xppf{\gamma}$ is $\gamma$-quantile of $\chi^2$-distribution with $1$ degree of freedom.
We note that if $\pmut{t+1}{n}$ does not change before and after the modification in \eqref{eq:clip_edgecase}, $[\m[t+1]]_{\jin_n}$ also does not change. Additionally, no changes are made to $\langle \A[t] \rangle_{\jin_n}$, namely,
\begin{align}
    \langle \A[t+1] \rangle_{\jin_n} \leftarrow \langle \A[t] \rangle_{\jin_n} \enspace.
\end{align}

\paragraph{Case of $~[\textnormal{\textsc{Enc}}(\m[t+1])]_{\jin_n} \in \{ z_{n,2}, \ldots, z_{n,L_n-1} \}$.}
Let us denote the nearest two thresholds to $[\m[t+1]]_{\jin_n}$ as follows:
\begin{align*}
    \llow{t+1}{n} &= \max \left\{ \ell_{n, l|l+1} \mid l \in \{1, \ldots, L_n-1\}, \ell_{n, l|l+1} < [\m[t+1]]_{\jin_n} \right\} \\
    \lup{t+1}{n} &= \min \left\{ \ell_{n, l|l+1} \mid l \in \{1, \ldots, L_n-1\}, [\m[t+1]]_{\jin_n} \leq \ell_{n, l|l+1} \right\}
\end{align*}
First, the marginal probabilities are calculated as
\begin{align*}
    \plow{t+1}{n} &= \Pr\left( [\bv]_{\jin_n} \leq \llow{t+1}{n} \right), \enspace \pup{t+1}{n} = \Pr\left( \lup{t+1}{n} < [\bv]_{\jin_n} \right) , \\
    \pmid{t+1}{n} &= 1 - \plow{t+1}{n} - \pup{t+1}{n} \enspace.
\end{align*}
Next, the marginal probabilities are modified so that $\plow{t+1}{n} \geq \alpha/2$, $\pup{t+1}{n} \geq \alpha/2$, and $\plow{t+1}{n} + \pup{t+1}{n} + \pmid{t+1}{n} = 1$ as follows:
\begin{align}
    \plow{t+1}{n} &\leftarrow \max\left\{\frac{\alpha}{2}, \plow{t+1}{n}\right\}, \quad \pup{t+1}{n} \leftarrow \max\left\{\frac{\alpha}{2}, \pup{t+1}{n}\right\} \label{eq:clip_plow} \\
    \Delt{t+1}{n} &\leftarrow \frac{1 - \plow{t+1}{n} - \pup{t+1}{n} - \pmid{t+1}{n}}{\plow{t+1}{n} + \pup{t+1}{n} + \pmid{t+1}{n} - 3\cdot \alpha/2} \label{eq:calc_Delta} \\
    \plow{t+1}{n} &\leftarrow \plow{t+1}{n} + \Delt{t+1}{n} (\plow{t+1}{n} - \alpha/2) \\
    \pup{t+1}{n} &\leftarrow \pup{t+1}{n} + \Delt{t+1}{n} (\pup{t+1}{n} - \alpha/2) \label{eq:clip_end}
\end{align}
Finally, $[\m[t+1]]_{\jin_n}$ and $\langle\A[t+1]\rangle_{\jin_n}$ are corrected as follows:
\begin{align*}
    [\m[t+1]]_{\jin_n} &\leftarrow \frac{\llow{t+1}{n} \sqrt{\Xppf{1-2\pup{t+1}{n}}} + \lup{t+1}{n} \sqrt{\Xppf{1-2\plow{t+1}{n}}}}{\sqrt{\Xppf{1-2\pup{t+1}{n}}} + \sqrt{\Xppf{1-2\plow{t+1}{n}}}} \\
    \langle\A[t+1]\rangle_{\jin_n} &\leftarrow \frac{\lup{t+1}{n} - \llow{t+1}{n}}{\sig[t+1]\sqrt{\langle\C[t+1]\rangle_{\jin_n}}\left(\sqrt{\Xppf{1-2\pup{t+1}{n}}} + \sqrt{\Xppf{1-2\plow{t+1}{n}}}\right)} 
\end{align*}
\subsection{CatCMA} \label{ssec:catcma}
CatCMA~\cite{CatCMA:2024} is a promising method for mixed-category black-box optimization.
CatCMA employs the joint probability distribution of multivariate Gaussian and categorical distributions.
To improve the search performance, the update rules of the CMA-ES and the Adaptive Stochastic Natural Gradient method~\cite{ASNG:2019} have been introduced into CatCMA.
The update rule for the multivariate Gaussian distribution in the joint probability distribution is essentially equivalent to that of the CMA-ES.
Therefore, this section mainly focuses on the update rule for the categorical distribution.

The family of categorical distributions on $\mathcal{C}$ is given as
\begin{align}
    p(\bc \mid \q) = \prod_{n=1}^{\Nca} \prod_{k=1}^{K_n} \q_{n,k}^{\bc_{n,k}} \enspace.
\end{align}
The distribution parameter $\q$ is defined as $\q=(\q_1^\top, \ldots, \q_\Nca^\top)^\top$ with
\begin{align}
    \q_n \in [0, 1]^{K_n} \enspace \text{s.t.} \enspace \sum_{k=1}^{K_n} \q_{n,k} = 1 \enspace,
\end{align}
where $\q_{n,k}$ denotes the probability of generating the $k$-th category in $n$-th dimension.

The natural gradient $G(\qt[t])$ is estimated by the Monte Carlo method with the samples and the evaluation of mixed-category solutions as
\begin{align}
    G(\qt[t]) = \sum_{i=1}^\lambda w_i (\bc_{i:\lambda} - \qt[t]) \enspace. \label{eq:CatCMA_ngr}
\end{align}
The distribution parameter is updated as
\begin{align}
    \qt[t+1] = \qt[t] + \delt[t] \frac{G(\qt[t])}{\left\| G(\qt[t]) \right\|_{F(\qt[t])}} \enspace, \label{eq:update_q}
\end{align}
where $\delt[t]$ is the trust region radius that is adapted in the following process, and $\left\| \cdot \right\|_{F(\qt[t])}$ denotes the Fisher norm relative to $\qt[t]$.
We introduce the accumulation of the estimated natural gradient as
\begin{align}
    \st[t+1] &= (1 - \beta)\st[t] + \sqrt{\beta (2 - \beta)} \left( F(\qt[t]) \right)^{\frac12} G(\qt[t]) \enspace, \label{eq:update_s} \\
    \gamt[t+1] &= (1 - \beta)^2 \gamt[t] + \beta(2 - \beta) \| G(\qt[t]) \|_{F(\qt)}^2 \enspace, \label{eq:update_gam}
\end{align}
where $\beta$ is set to $\delt[t] / \bigl( \sum_{n=1}^{\Nca} (K_n - 1) \bigl)$. The trust region radius $\delt[t]$ is adapted as
\begin{align}
    \delt[t+1] = \delt[t] \exp \left( \beta \left( \frac{\| \st[t+1] \|^2}{\alpha_{\mathrm{snr}}} - \gamt[t+1] \right) \right) \enspace, \label{eq:update_del}
\end{align}
where $\alpha_{\mathrm{snr}}$ is a constant.

To prevent numerical errors due to the excessive convergence of the multivariate Gaussian distribution, $\sig[t+1]$ is modified as
\begin{align}
    \sig[t+1] \leftarrow \max \left\{ \sig[t+1],  \sqrt{\frac{\Lambda^{\min}}{ \min\{\mathrm{eig}(\C[t+1])\} }} \right\} \enspace, \label{eq:post-process_sigma}
\end{align}
where $\textrm{eig}(\C[t+1])$ is the set of the eigenvalues of $\C[t+1]$, and $\Lambda^{\min}$ is the lower bound of the eigenvalues of the multivariate Gaussian distribution.

When some elements of $\q$ become too small, the corresponding categories are hardly generated.
To prevent this, $\q$ is modified by the margin correction as follows:
\begin{align}
    &\qt[t+1]_{n,k} \leftarrow \max \left\{ \qt[t+1]_{n,k}, \qmin_n \right\} \label{eq:margin_correction_1} \\
    &\qt[t+1]_{n,k} \leftarrow \qt[t+1]_{n,k} + \frac{1 - \sum_{k'=1}^{K_n} \qt[t+1]_{n,k'} }{\sum_{k'=1}^{K_n} (\qt[t+1]_{n,k'} - \qmin_n)} (\qt[t+1]_{n,k} - \qmin_n) \label{eq:margin_correction_2}
\end{align}
This margin correction guarantees $\qt[t+1]_{n,k} \geq \qmin_n$ and $\sum_{k=1}^{K_n} \qt[t+1]_{n,k}= 1$.

\section{Review of Existing Integer Handling}
The goal of our study is to develop an MV-BBO method that handles continuous, integer, and categorical variables by introducing an integer handling into the MC-BBO method, CatCMA.
However, it is not clear whether existing integer handling methods for MI-BBO are effective for the simultaneous optimization of these three types of variables.
This section reviews existing integer handling methods to propose a new integer handling suitable for CatCMA in the next section.

\begin{figure}[t]
    \centering    \includegraphics[width=0.6\linewidth]{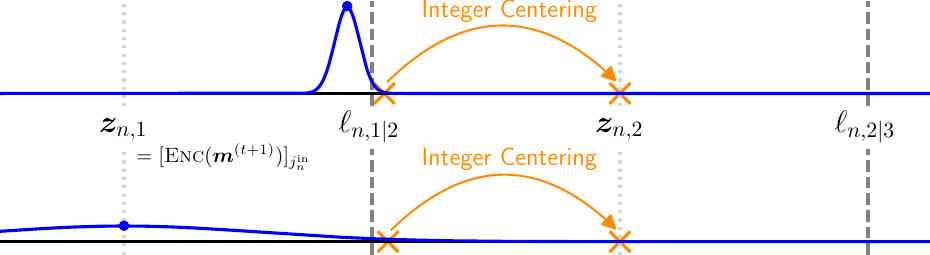}
    \caption{Integer centering under multivariate Gaussian distributions with different standard deviations.}
    \label{fig:centering}
\end{figure}

\subsection{Existing Integer Handling for CMA-ES} \label{ssec:existing_integer_handlings}
This subsection summarizes existing integer handling for the CMA-ES. Based on this summary, we discuss the requirements for introducing integer handling to CatCMA.

\paragraph{CMA-ES with Integer Mutation}
\citet{hansen_cma-es_2011} proposed \textit{integer mutation} as an integer handling for the CMA-ES.
Integer mutation randomly mutates the elements of an integer variable to one of its nearby integers if it is determined that the elements of a certain dimension of the integer variable are fixed.
Hansen reported that it is not suitable for binary variables or $k$-ary integers in $k < 10$.

\paragraph{CMA-ES with Margin}
As introduced in Section~\ref{ssec:cmawm}, the CMA-ESwM is a variant of the CMA-ES that introduces the margin correction.
The margin correction is originally a common technique in estimation-of-distribution algorithms to prevent generated binary variables from being fixed at 0 or 1 by correcting their generation probabilities to a certain range.
The CMA-ESwM, which was inspired by this and introduced a lower bound for the marginal probability, can also perform efficient optimization for objective functions that include binary variables.

\paragraph{LB+IC-CMA-ES}
\citet{LB+IC-CMA-ES:2024} proposed two simple modifications of the CMA-ES to handle MI-BBO: the lower bound for the standard deviation of integer coordinates and \textit{integer centering}.
This lower bound is intended to prevent the \emph{mutation rate}, the proportion of candidate solutions that are on a different integer plateau than the mean vector, from becoming too small in a similar way to margin correction.
In order for the multivariate Gaussian distribution to move to the different integer plateau, it is necessary to generate superior samples outside of the plateau with the mean vector.
Marty et al. call such a generation of samples \textit{successful integer mutation}.
They also pointed out that a successful integer mutation often has little effect on the update of the multivariate Gaussian distribution if its value is close to a discretization threshold.
To address this issue, integer centering sets successfully mutated values to the center of the integer interval, as shown in Figure~\ref{fig:centering}.
While the appropriate setting of the lower bound for the standard deviation of integer coordinates has been thoroughly analyzed, it is important to note that this does not bound the mutation rate of integers to a specific value.

\paragraph{Requirements for Integrating into CatCMA}
As with these integer handling methods, CatCMA also introduces the margin correction shown in \eqref{eq:margin_correction_1} and \eqref{eq:margin_correction_2} to prevent the fixation of categorical variables.
The promising margin value $\qmin_n$ is analytically derived considering the trade-off between the risk of fixation and the impact of non-optimal categories on the sample ranking.
In order not to violate this design principle, we propose to incorporate the margin correction in the CMA-ESwM into CatCMA.
This is because the margin correction can bound the mutation rate of integers by a certain value, and it allows an explicit evaluation of the impact of mutated integers on the sample ranking.

\subsection{Assessment of Assumptions for Derivation of Promising Margin} \label{ssec:assessment_assumptions}
In the previous subsection, we proposed to integrate the margin correction into CatCMA.
Similar to derivation of the promising margin $\qmin_n$ in CatCMA, we aim to analytically derive the promising margin $\alpha$, whose recommended value has been determined experimentally in~\cite{CMAwM:2022}.
However, we will experimentally demonstrate that the assumptions required for the analytical derivation of the promising margin $\alpha$ are not satisfied by the CMA-ESwM.

In \cite[Proposition~4.1]{CatCMA:2024}, which derives the promising margin $\qmin_n$ for CatCMA, it is assumed that the categorical distribution has converged to the optimal state.
This assumption is imposed to evaluate the number of samples that contain non-optimal categories due to the margin, which is satisfied as long as such samples are not used in the update.
On the other hand, it is not reasonable to assume that the part of the multivariate Gaussian distribution responsible for generating integer variables has converged to the optimal state, which will be experimentally demonstrated in the following.

\paragraph{Experimental Settings.}
We used \textsc{EllipsoidInt} defined as follows:
\begin{align}
    f(\x, \z) = \sum_{n=1}^{\Nco} 10^{6\frac{n-1}{\Nco+\Nin-1}} \x_n^2 + \sum_{n=1}^{\Nin} 10^{6\frac{\Nco+n-1}{\Nco+\Nin-1}}\z_n^2
\end{align}
The number of dimensions for continuous and integer variables was set to $3$. The integer space was set as $\mathcal{Z}_1 = \mathcal{Z}_2 = \mathcal{Z}_3 = \{-10, -9, \ldots, 9, 10\}$.
The initial parameters of the multivariate Gaussian distribution were given as $\m[0] = (3,\ldots,3)$, $\C[0] = \mathbf{I}_{\Nmi}$, and $\sig[0] = 3$.
The margin $\alpha$ was set to the recommended value of $1/(\lambda \Nmi)$ in the CMA-ESwM, and to a larger value of $1 - 0.73^{1/\Nin}$, which is the setting derived in the next section.
The weights $w$ were used in two ways: one including negative weights, as used in the LB+IC-CMA-ES and the CMA-ESwM, and the other with non-negative weights used in CatCMA.
CatCMA assumes that the impact of the samples in the lower half of the ranking is ignored.
Thus, we also experiment with non-negative weights to discuss the impact of the samples in the lower half of the ranking.
In the LB+IC-CMA-ES, we introduced box constraint~\cite{BoxConst:2009} to handle edge integers, as in~\cite{LB+IC-CMA-ES:2024}.
The other hyperparameters of the LB+IC-CMA-ES and the CMA-ESwM were set to their recommended values.

\begin{figure*}[t]
    \centering
    \includegraphics[width=0.99\linewidth]{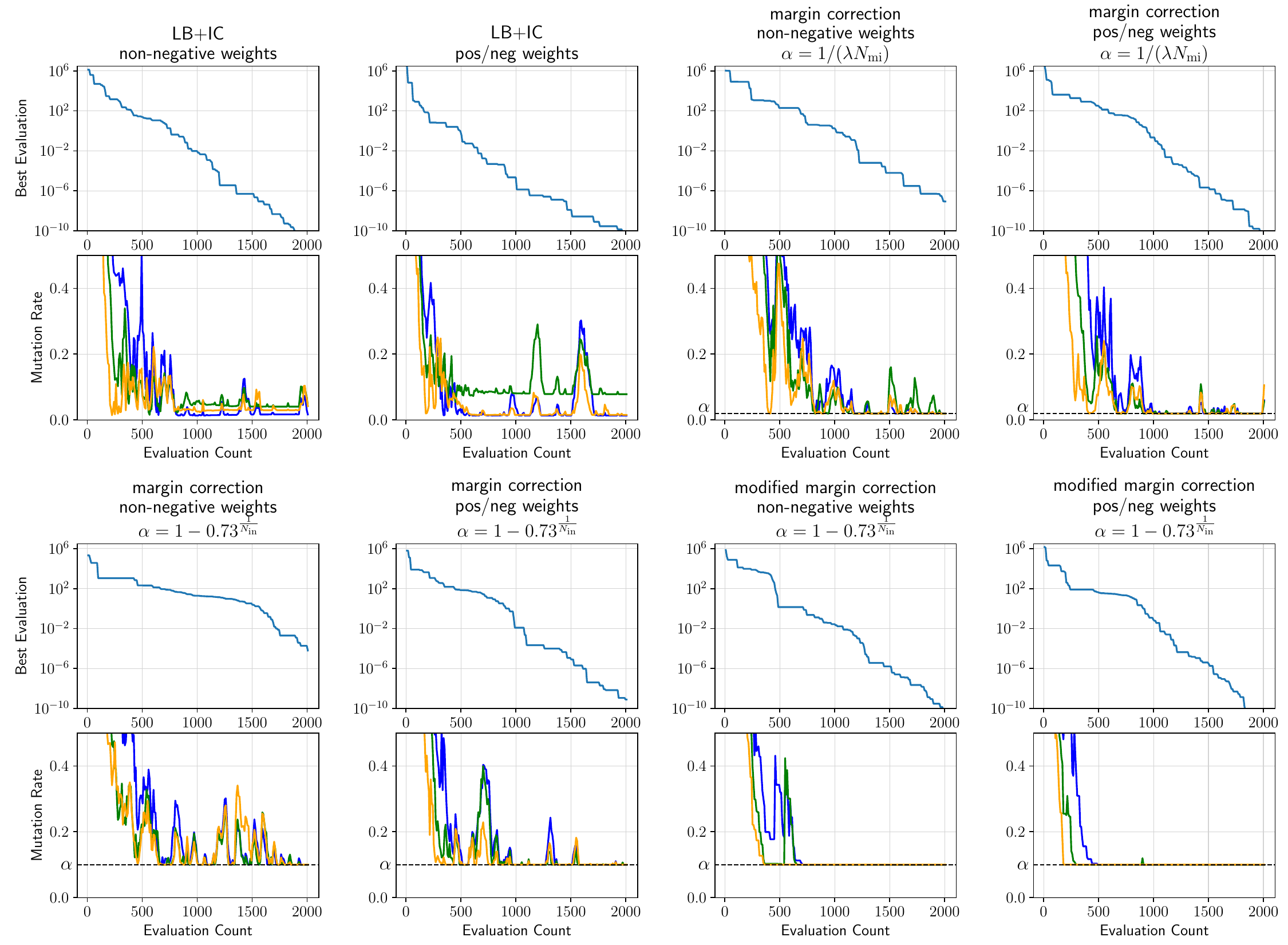}
    \caption{The best evaluation value and the mutation rate for each dimension in a typical run on \textsc{EllipsoidInt}.}
    \label{fig:margin_typical}
\end{figure*}

\paragraph{Results.}
Figure~\ref{fig:margin_typical} shows the results of a typical trial for each setting in \textsc{EllipsoidInt}, including the results discussed in Section~\ref{ssec:derivation_margin}.
The value of $p_{\mathrm{mut},n}^{\smash{(t)}}$ or $p_{\mathrm{low},n}^{\smash{(t)}} + p_{\mathrm{up},n}^{\smash{(t)}}$ as the mutation rate is plotted in the even row.
When the part of the multivariate Gaussian distribution responsible for generating integer variables has converged to the optimal state, these mutation rates are expected to remain equal to the margin value $\alpha$ shown by the dashed line in Figure~\ref{fig:margin_typical}.
However, we observe that the values of these mutation rates fluctuate even when the integer variables are sufficiently optimized.
This is because when optimizing continuous variables, the step-size increases or decreases, which also affects the integer coordinates of the multivariate Gaussian distribution.
Therefore, deriving the promising margin $\alpha$ based on the assumption discussed above can lead to underestimating the probability of generating candidate solutions that include non-optimal integer variables.

We also discuss that this fluctuation is not only inconvenient for deriving the promising margin, but also undesirable from the perspective of efficient optimization.
The method of adapting mutation rate, such as $1/5$-success rule~\cite{Rechenberg:1973}, is designed to increase the mutation rate if the mutation is successful and decrease it if it is not.
On the other hand, the CMA-ESwM increases the mutation rate of integers even if the mutation of integer variables is not successful at the end of the optimization process.
As a result, the probability of generating non-optimal integers increases unintentionally, which has a negative impact on the optimization of continuous variables.

\section{Proposed Integer Handling} \label{sec:proposed_integer_handling}
In this section, we propose a new integer handling suitable for integrating into CatCMA based on the previous section.

\subsection{Modified Margin Correction}
We propose a mechanism to suppress unnecessary fluctuations of the mutation rate.
The proposed mechanism imposes the upper bound of the mutation rate depending on whether a successful integer mutation occurs.
This upper bound is motivated by the analytical derivation of a promising margin and the perspective of adapting the mutation rate for efficient optimization.

Based on \cite{LB+IC-CMA-ES:2024}, we consider that a successful integer mutation occurs in the dimension $n \in \{1, \ldots, \Nin\}$ when the following condition is satisfied:
\begin{align}
    \exists i \in \{1\!:\!\lambda, \ldots, \mu\!:\!\lambda\} \enspace \text{s.t.} \enspace [\z_i]_{n} \neq [\Enc(\m[t])]_{\jin_n}
\end{align}
To uniformly refer to the mutation rate in the previous iteration as $\pmut{t}{n}$, we insert the following update after \eqref{eq:clip_end}:
\begin{align}
    \pmut{t+1}{n} \leftarrow \plow{t+1}{n} + \pup{t+1}{n} 
\end{align}
For dimensions where a successful integer mutation occurs, the margin correction in Section~\ref{sssec:margin_correction} is performed, and for other dimensions, the following modified margin correction is performed.

\paragraph{Case of $~[\textnormal{\textsc{Enc}}(\m[t+1])]_{\jin_n} \in \{ z_{n,1}, z_{n,L_n} \}$.}
The update of \eqref{eq:clip_edgecase} is modified to the following:
\begin{align}
    \pmut{t+1}{n} \leftarrow \max\left\{ \alpha, \min\left\{ \pmut{t+1}{n}, \pmut{t}{n} \right\} \right\} \label{eq:clip_pmut_modified}
\end{align}

\paragraph{Case of $~[\textnormal{\textsc{Enc}}(\m[t+1])]_{\jin_n} \in \{ z_{n,2}, \ldots, z_{n,L_n-1} \}$.}
The updates \eqref{eq:clip_plow} to \eqref{eq:calc_Delta} are modified to the following:
\begin{align}
    \plow{t+1}{n} &\leftarrow \max\left\{\frac{\alpha}{2}, \plow{t+1}{n}\right\}, \quad \pup{t+1}{n} \leftarrow \max\left\{\frac{\alpha}{2}, \pup{t+1}{n}\right\} \label{eq:clip_plow_modified} \\
    \pmid{t+1}{n} &\leftarrow \max\left\{1 - \pmut{t}{n}, \pmid{t+1}{n}\right\} \label{eq:clip_pmid_modified} \\
    \Delt{t+1}{n} &\leftarrow \frac{1 - \plow{t+1}{n} - \pup{t+1}{n} - \pmid{t+1}{n}}{\plow{t+1}{n} + \pup{t+1}{n} + \pmid{t+1}{n} - \alpha - (1 - \pmut{t}{n})} \label{eq:calc_Delta_modified}
\end{align}
The above modification ensures $\plow{t+1}{n} \geq \alpha / 2$, $\pup{t+1}{n} \geq \alpha / 2$, $\plow{t+1}{n} + \pup{t+1}{n} \leq \pmut{t}{n}$, and $\plow{t+1}{n} + \pup{t+1}{n} + \pmid{t+1}{n} = 1$.


\subsection{Introduction of Integer Centering} \label{ssec:integer_centering}
To further facilitate the escape of the multivariate Gaussian distribution from a plateau, we introduce integer centering proposed in \cite{LB+IC-CMA-ES:2024}.
Integer centering can be implemented independently of the margin correction, since it corrects $\bv_{1:\lambda}, \ldots, \bv_{\mu:\lambda}$ and $\y_{1:\lambda}, \ldots, \y_{\mu:\lambda}$ after evaluation by the objective function.
However, naively introducing integer centering into the CMA-ESwM can lead to unstable optimization in some cases.
As shown in the top of Figure~\ref{fig:centering}, when the mean vector stays in the region where edge integers are generated, the corresponding standard deviation can remain very small.
This is because only the mean vector is corrected in the case of edge integers.
In such cases, since the movement of the samples by integer centering becomes very large compared to the standard deviation of the multivariate Gaussian distribution, the evolution path $\ps[t]$ rapidly becomes long.
To address this problem, we introduce the following correction to $\langle \A[t+1] \rangle_{\jin_n}$ so that the standard deviation of the multivariate Gaussian distribution does not become too small even in the case of edge integers.
\begin{align}
    \langle \A[t+1] \rangle_{\jin_n} \leftarrow \max  \left\{  \frac{\left| [\Enc(\m[t+1])]_{\jin_n} - \lnei{t+1}{n} \right|}{\sig[t+1] \sqrt{\langle \C[t+1] \rangle_{\jin_n}\Xppf{1-2\alpha}}},   \langle \A[t] \rangle_{\jin_n} \right\} \label{eq:correction_A_new}
\end{align}
After this correction, $[\m[t+1]]_{\jin_n}$ is corrected using $\pmut{t+1}{n}$ previously calculated in \eqref{eq:clip_pmut_modified} as follows:
\begin{align}
    \begin{split}
        &[\m[t+1]]_{\jin_n} \leftarrow \rev{\lnei{t+1}{n}} + \sign \left( [\m[t+1]]_{\jin_n} - \rev{\lnei{t+1}{n}} \right) \\
        &\qquad\qquad\qquad\qquad\qquad \cdot \sig[t+1] \langle \A[t+1] \rangle_{\jin_n} \sqrt{\langle \C[t+1] \rangle_{\jin_n}\Xppf{1-2\pmut{t+1}{n}}}
    \end{split}
\end{align}
When $\langle \A[t+1] \rangle_{\jin_n}$ is corrected to the lower bound in \eqref{eq:correction_A_new} and $\pmut{t+1}{n}$ is $\alpha$, $[\m[t+1]]_{\jin_n}$ is corrected to match $\z_{n,1}$ or $\z_{n,L_n}$, as shown in the bottom of Figure~\ref{fig:centering}.
A summary of the modified margin correction is provided in Appendix~\ref{apdx:modified-margin-correction} to facilitate following the revised and added update rules.

\subsection{Derivation of Promising Margin Settings} \label{ssec:derivation_margin}
Before the derivation of promising margin settings, we confirm that the modified margin correction can suppress unnecessary fluctuations of the mutation rate.
The results of a typical trial introducing the modified margin correction and the integer centering are included in Figure~\ref{fig:margin_typical}.
Since no upper bound is imposed on the mutation rate in the case of a successful integer mutation, the impact on integer variable optimization is minimal.
Moreover, when the optimization of integer variables has been sufficiently performed, a successful integer mutation does not occur, and the upper bound prevents the mutation rate from fluctuating.
Therefore, we can reasonably assume the convergence of the mutation rate for deriving promising margin settings.

Then, similar to \cite[Proposition~4.1]{CatCMA:2024}, we can derive the margin $\alpha$ that sufficiently reduces the impact of non-optimal integers caused by the margin on the solution ranking.
\begin{prop}
    \label{prop:int_margin}
    Let $\z^\opt \in \mathcal{Z}$ be an optimal integer variable. Assume that the parameters of the multivariate Gaussian distribution satisfy the following:
    \begin{align}
        [\textnormal{\textsc{Enc}}(\m[t])]_{\jin_n} &= \z^\opt_n \enspace \text{for all} \enspace n \in \{ 1, \ldots, \Nin \} \label{eq:int_prop_cond1} \\
        \pmut{t}{n} &= \alpha \enspace \text{for all} \enspace n \in \{ 1, \ldots, \Nin \} \label{eq:int_prop_cond2}
    \end{align}
    Let $\lambda_\mathrm{non}$ be the random variable that counts the number of samples containing non-optimal integer variables among the $\lambda$ samples in the $t$-th iteration of the CMA-ESwM. If the margin satisfies
    \begin{align}
        \alpha = 1 - 0.73^\frac{1}{\Nin} \enspace, \label{eq:promising_margin_int}
    \end{align}
    it holds, for any $\lambda \geq 6$,
    \begin{align}
        \Pr\left( \lambda_\non \leq \lambda - \left\lfloor \frac{\lambda}{2} \right\rfloor \right) \geq 0.95 \enspace. \label{eq:lambda_non_prob_int}
    \end{align}
\end{prop}
The proof of Proposition~\ref{prop:int_margin} can be found in Appendix~\ref{apdx:proof-int-margin}.
This proposition provides the margin that ensures that the top $\mu = \lfloor \lambda/2 \rfloor$ solutions do not contain non-optimal integers with high probability when the part of the multivariate Gaussian distribution responsible for generating integer variables has converged to the optimal state.
This implies that, in the updates \eqref{eq:update_m} to \eqref{eq:update_pc}, the solutions containing non-optimal integers are almost ignored.
On the other hand, when using negative weights, the update of the covariance matrix \eqref{eq:update_C} is affected by such samples, which will be investigated in the next subsection. 

\begin{figure*}[t]
    \centering
    \includegraphics[width=0.99\linewidth]{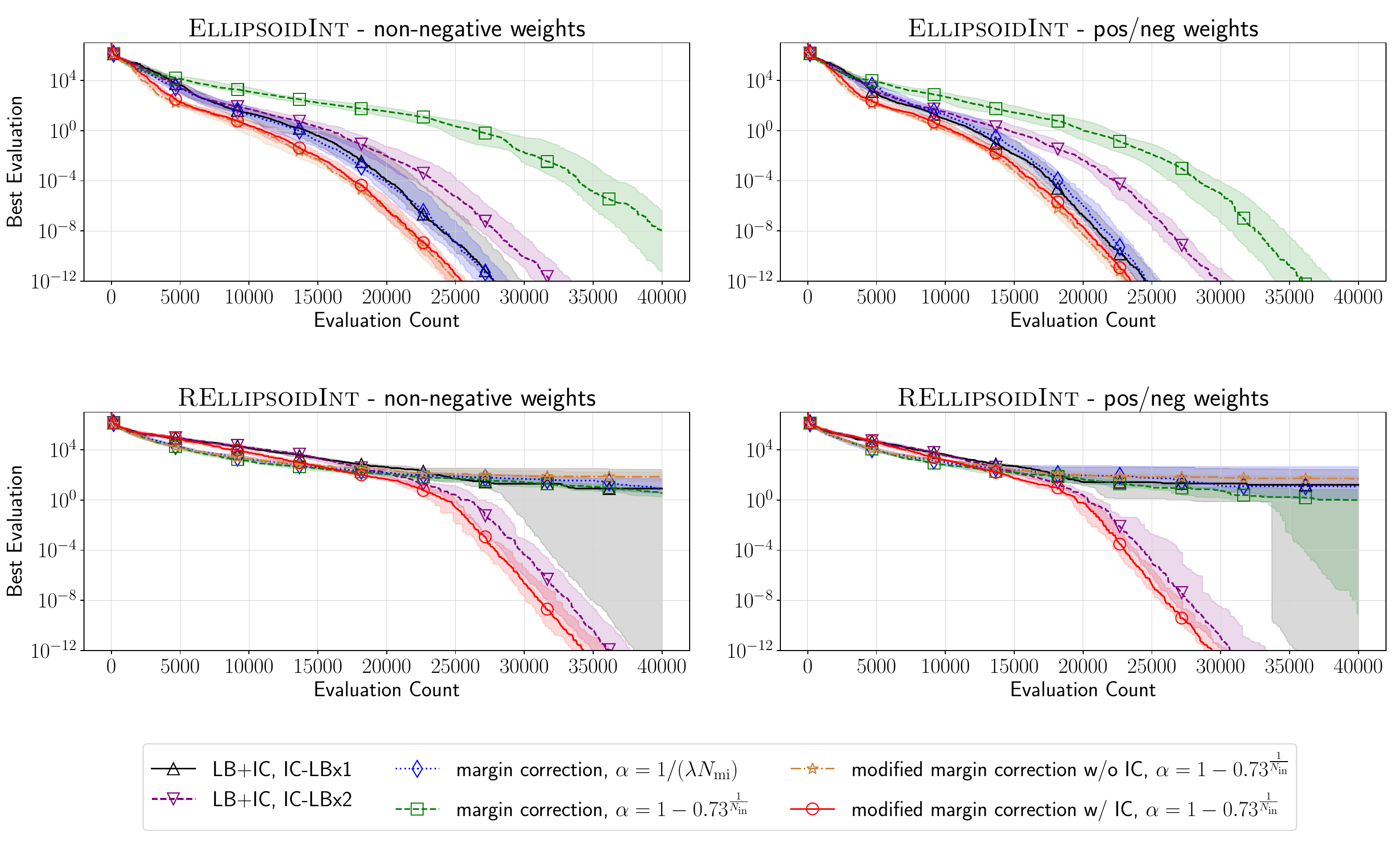}
    \vspace{-2mm}
    \caption{
    The medians and inter-quartile ranges of the best evaluation values over 20 independent trials.
    }
    \label{fig:margin_compe}
\end{figure*}

\subsection{Experimental Evaluation of Proposed Margin Correction} \label{ssec:modified_margin}
This subsection evaluates the modified margin correction and confirms its effectiveness.
We also discuss the appropriate weights, addressing the question raised in the previous section.

\paragraph{Experimental Settings.}
In addition to \textsc{EllipsoidInt} in Section~\ref{ssec:assessment_assumptions}, we used \textsc{ReversedEllipsoidInt} (\textsc{REllipsoidInt}) defined as follows:
\begin{align}
    f(\x, \z) = \sum_{n=1}^{\Nco} 10^{6\frac{\Nin+n-1}{\Nco+\Nin-1}} \x_n^2 + \sum_{n=1}^{\Nin} 10^{6\frac{n-1}{\Nco+\Nin-1}}\z_n^2
\end{align}
In \textsc{EllipsoidInt}, integer variables have a greater impact on the evaluation value than continuous variables, whereas in \textsc{REllipsoidInt}, continuous variables have a greater impact than integer variables.
The numbers of dimensions for both continuous and integer variables were set to $20$. The integer space was set as $\mathcal{Z}_1 = \cdots = \mathcal{Z}_{20} = \{-10, -9, \ldots, 9, 10\}$.
The initial parameters of the multivariate Gaussian distribution were given as $\C[0] = \mathbf{I}_{\Nmi}$ and $\sig[0] = 1$, and $\m[0]$ was uniformly sampled in $[1,3]^{\Nmi}$.
In the LB+IC-CMA-ES, we used the settings of \textsf{IC-LBx1} and \textsf{IC-LBx2} in~\cite{LB+IC-CMA-ES:2024}.
The other settings were the same as those in Section~\ref{ssec:assessment_assumptions}.

\paragraph{Results.}
Figure~\ref{fig:margin_compe} shows the best evaluation values for each setting.
Overall, we can see that using negative weights leads to more efficient optimization compared to using non-negative weights.
The covariance matrix adaptation using negative weights is called \textit{active covariance matrix adaptation}~\cite{ACMA:2006}, and it allows for efficient adaptation.
Although the use of negative weights cannot ignore non-optimal integers in the covariance matrix adaptation, the contribution of the active covariance matrix adaptation improves performance overall.

In \textsc{EllipsoidInt}, the modified margin correction performs better than the original margin correction.
Even with the margin setting of $\alpha=1-0.73^{1/\Nin}$, the original margin correction does not satisfy the assumption discussed in Section~\ref{ssec:assessment_assumptions}, leading to higher mutation rates and inferior performance.
In \textsc{REllipsoidInt}, the modified margin correction without integer centering slightly increases the fixation of integers due to the upper bound of the mutation rate.
Notably, this fixation is resolved by integer centering.
The proposed integer handling performs best on the two functions with different properties without hyperparameter tuning.
\begin{algorithm}[t!]
    \caption{CatCMA with Margin}
    \begin{algorithmic}[1]
        \REQUIRE $\m[0], \sig[0], \C[0], \qt[0]$
        \STATE $\A[0] = \mathbf{I}_\Nmi$, $\ps[0] = \pc[0] = \boldsymbol{0}$, $\delt[0] = 1$, $\st[0] = \boldsymbol{0}$, $\gamt[0] = 0$, $\pmut{0}{1}=\cdots=\pmut{0}{\Nin} = 1$, $t=0$
        \WHILE{termination conditions are not met}
        \FOR{$i=1,\ldots,\lambda$}
            \STATE Sample $\bxi_i \sim \mathcal{N}(\boldsymbol{0}, \mathbf{I}_\Nmi)$
            \STATE Sample $\bc_i \sim p(\bc \mid \qt[t])$
            \STATE $\y_i \leftarrow \sqrt{\C[t]} \bxi_i$
            \STATE $\bv_i \leftarrow \m[t] + \sig[t] \A[t] \y_i$
            \FOR{$n=1,\ldots,\Nco$}
                \STATE $[\x_i]_n \leftarrow [\Enc(\bv_i)]_{j^{\mathrm{co}}_n}$
            \ENDFOR
            \FOR{$n=1,\ldots,\Nin$}
                \STATE $[\z_i]_n \leftarrow [\Enc(\bv_i)]_{j^{\mathrm{in}}_n}$
            \ENDFOR
        \ENDFOR
        \STATE Evaluate $(\x_1, \z_1, \bc_1), \ldots, (\x_\lambda, \z_\lambda, \bc_\lambda)$ on $f(\x, \z, \bc)$
        \STATE Correct $\bv_{1:\lambda},\ldots,\bv_{\mu:\lambda}$ and $\y_{1:\lambda},\ldots,\y_{\mu:\lambda}$ by integer centering in \cite{LB+IC-CMA-ES:2024}
        \STATE Update $\m[t]$, $\ps[t]$, $\pc[t]$, $\C[t]$, and $\sig[t]$ according to Section~\ref{sssec:update_based_on_cmaes}
        \STATE Update $\qt[t]$, $\st[t]$, $\gamt[t]$, $\delt[t]$, and modify $\sig[t+1]$ and $\qt[t+1]$ according to Section~\ref{ssec:catcma}
        \STATE Update $\m[t+1]$, $\A[t]$, $\pmut{t}{1}, \ldots, \pmut{t}{\Nin}$ by the modified margin correction
        \STATE $t \leftarrow t+1$
        \ENDWHILE
    \end{algorithmic}
    \label{alg:CatCMAwM}
\end{algorithm}
\section{Proposed Method: CatCMA with Margin}
In this section, we propose CatCMA with Margin (CatCMAwM), an MV-BBO method that can handle continuous, integer, and categorical variables.
As shown in Algorithm~\ref{alg:CatCMAwM}, CatCMAwM is developed by incorporating the integer handling proposed in the previous section into CatCMA.
As a refinement, we modify the estimation in \eqref{eq:CatCMA_ngr} to use the $\mu$-best solutions for the following reason: to use negative weights solely for active covariance matrix adaptation, we align the weights in the estimation of \eqref{eq:CatCMA_ngr} in CatCMA with those used in the update of the mean vector in \eqref{eq:update_m}, as in \cite{CatCMA:2024}.
We can derive the promising margin values $\alpha$ and $\qmin_n$ in CatCMAwM similar to Proposition~\ref{prop:int_margin} and \cite[Proposition~4.1]{CatCMA:2024}.

\begin{prop}
    \label{prop:mix_margin}
    Let $\z^\opt \in \mathcal{Z}$ be an optimal integer variable. Assume that the parameters of the multivariate Gaussian distribution satisfy the following:
    \begin{align}
        [\textnormal{\textsc{Enc}}(\m[t])]_{\jin_n} &= \z^\opt_n \enspace \text{for all} \enspace n \in \{ 1, \ldots, \Nin \} \label{eq:mix_prop_cond1} \\
        \pmut{t}{n} &= \alpha \enspace \text{for all} \enspace n \in \{ 1, \ldots, \Nin \} \label{eq:mix_prop_cond2}
    \end{align}
    Without loss of generality, assume that categories of the optimal solution are the first categories in all dimensions. Assume that the parameter of the categorical distribution satisfies
    \begin{align}
        \begin{split}
            &\qt[t]_{n,1} = 1 - \qmin_n (K_n - 1) \enspace, \\
            &\qt[t]_{n, k} = \qmin_n \quad \text{for all} \enspace k \in \{2, \ldots, K_n \}
        \end{split} \label{eq:assum_cat}
    \end{align}
    for all $n \in \{ 1, \ldots, \Nca \}$.
    Let $\lambda_\mathrm{non}$ be the random variable that counts the number of samples containing non-optimal integer and/or categorical variables among the $\lambda$ samples in the $t$-th iteration of CatCMAwM. If the margins satisfy
    \begin{align}
        \alpha &= 1 - 0.73^\frac{1}{\Nin+\Nca} \enspace, \label{eq:promising_margin_alpha} \\
        \qmin_n &= \frac{1}{K_n - 1}\left( 1 - 0.73^\frac{1}{\Nin+\Nca} \right) \enspace \text{for all} \enspace n \in \{ 1, \ldots, \Nca \} \enspace, \label{eq:promising_margin_qmin}
    \end{align}
    it holds, for any $\lambda \geq 6$,
    \begin{align}
        \Pr\left( \lambda_\non \leq \lambda - \left\lfloor \frac{\lambda}{2} \right\rfloor \right) \geq 0.95 \enspace. \label{eq:lambda_non_prob_mix}
    \end{align}
\end{prop}
The proof of Proposition~\ref{prop:mix_margin} can be found in Appendix~\ref{apdx:proof-mix-margin}.
This proposition focuses on the phase where optimization of both integer and categorical variables has been sufficiently performed.
This margin setting is expected to achieve a balance between avoiding the fixation of discrete variables and efficient continuous variable optimization.

The code for CatCMAwM will be made available at \textcolor{blue}{\url{https://github.com/CyberAgentAILab/cmaes}}~\cite{nomura2024cmaes}.
\section{Experiments on Single-Objective Optimization Problems} \label{sec:experiments-single-objective}
This section evaluates the search performance of CatCMAwM on single-objective MV-BBO problems.
We used the following benchmark functions.
\begin{itemize}
    \item SphereIntCategoricalOneMax (\textsc{SphereIntCOM})
    \begin{align}
        f(\x, \z, \bc) = \sum_{n=1}^{\Nco} \x_n^2 + \sum_{n=1}^{\Nin} \z_n^2 + \Nca - \sum_{n=1}^{\Nca} \bc_{n,1}
    \end{align}
    \item EllipsoidIntCategoricalLeadingOnes (\textsc{EllipsoidIntCLO})
    \begin{align}
        f(\x, \z, \bc) = \sum_{n=1}^{\Nco} 10^{6\frac{n-1}{\Nco+\Nin-1}} \x_n^2 + \sum_{n=1}^{\Nin} 10^{6\frac{\Nco+n-1}{\Nco+\Nin-1}}\z_n^2 + \Nca - \sum_{n=1}^{\Nca} \prod_{n'=1}^n \bc_{n',1}
    \end{align}
    \item ReversedEllipsoidIntCategoricalLeadingOnes (\textsc{REllipsoidIntCLO})
    \begin{align}
        f(\x, \z, \bc) = \sum_{n=1}^{\Nco} 10^{6\frac{\Nin + n-1}{\Nco+\Nin-1}} \x_n^2 + \sum_{n=1}^{\Nin} 10^{6\frac{n-1}{\Nco+\Nin-1}}\z_n^2 + \Nca - \sum_{n=1}^{\Nca} \prod_{n'=1}^n \bc_{n',1}
    \end{align}
    \item MixedVariableProximity (\textsc{MVProximity})
    \begin{align}
        \begin{split}
            f(\x, \z, \bc) = \sum_{n=1}^{\Nco (= \Nca)} \left( \frac{\x_n}{x_{\scale}} - \boldsymbol{\zeta}_n \right)^2 &+ \sum_{n=1}^{\Nin (= \Nca)} \left( \frac{\z_n}{z_{\scale}} - \boldsymbol{\zeta}_n \right)^2 + \sum_{n=1}^{\Nca} \boldsymbol{\zeta}_n \enspace, \\
            &\qquad\qquad \text{where} \enspace \boldsymbol{\zeta}_n = \frac{1}{K_n}\sum_{k=1}^{K_n} (k-1) \bc_{n,k} \enspace.
        \end{split}
    \end{align}
\end{itemize}
The first three functions are the sum of existing mixed-integer and categorical benchmark functions.
In \textsc{MVProximity}, there are dependencies between continuous and categorical variables, and between integer and categorical variables.

\paragraph{Experimental Settings}
The integer space was set as $\mathcal{Z}_1 = \cdots = \mathcal{Z}_{\Nin} = \{-3, -2, \ldots, 2, 3\}$.
The numbers of categories were set to $5$.
In \textsc{MVProximity}, we set $x_{\smash{\scale}}$ and $z_{\smash{\scale}}$ to $3$ according to the search space.
As baseline state-of-the-art methods, we experimented with CASMOPOLITAN~\cite{casmopolitan:2021}, SMAC3~\cite{SMAC:2011, SMAC3:2022}, and Tree-structured Parzen Estimator (TPE)~\cite{TPE:2011} implemented by Optuna~\cite{optuna:akiba2019}.
In the three methods, the ranges of the continuous variables were set to $[-3, 3]$. The evaluation budget on the CASMOPOLITAN was set to $400$ due to long internal computation time.
To confirm the effectiveness of the proposed integer handling, we also experimented with two CatCMA methods naively incorporating the default margin correction and the integer handling of LB+IC-CMA-ES, using the default hyperparameters such as the margin value for each method.
In the CatCMA-based methods, the initial distribution parameters were given as $\C[0] = \mathbf{I}_{\Nmi}$ and $\sig[0] = 1$, $\boldsymbol{q}_n^{\smash{(0)}} = 1/K_n$ and $\m[0]$ was uniformly sampled in $[1,3]^{\smash{\Nmi}}$, and sample size was set as $\lambda = 4 + \lfloor 3 \ln (\Nco + \Nin + \Nca) \rfloor$.

\begin{figure}[t]
    \centering
    \includegraphics[width=0.99\linewidth]{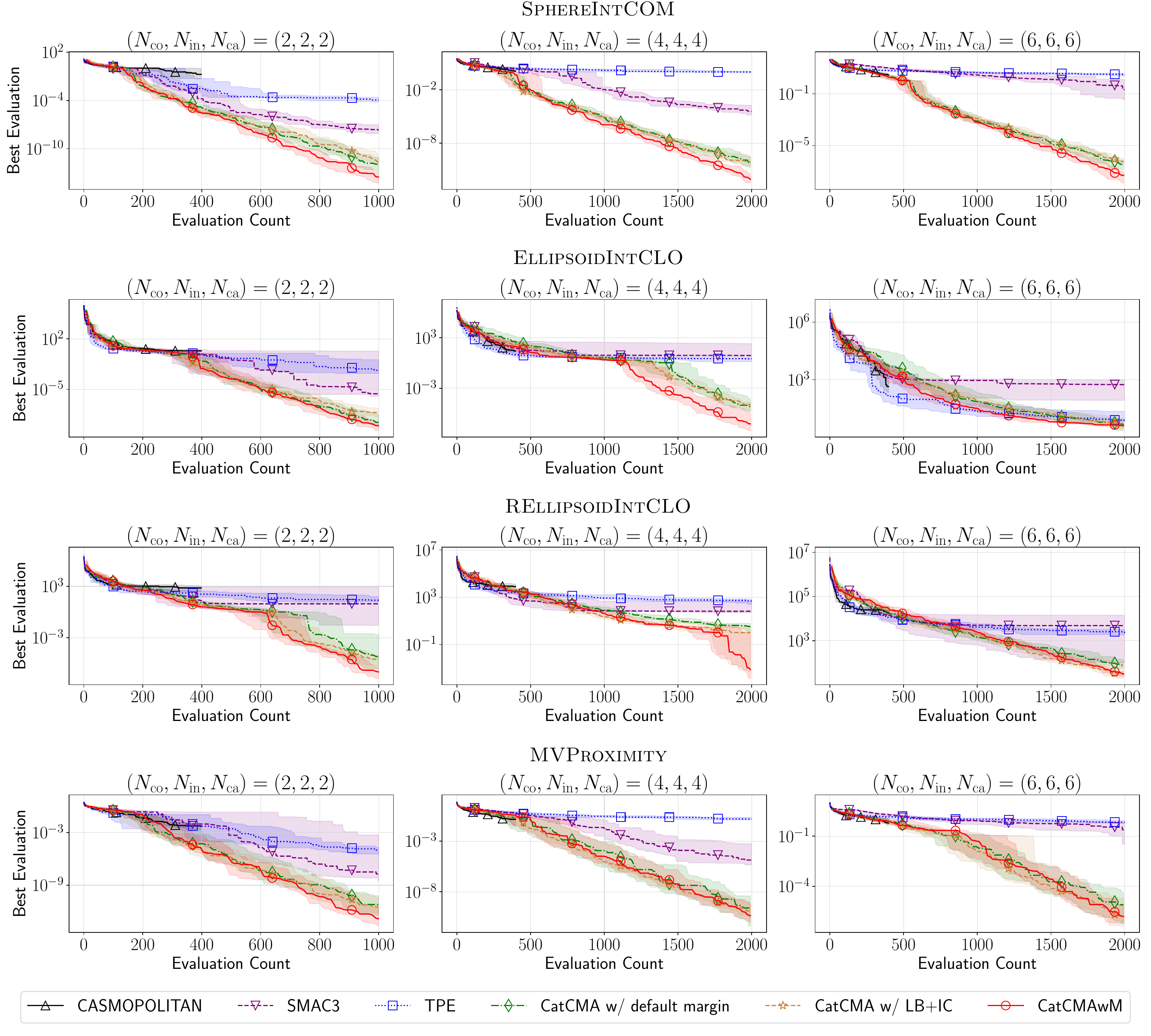}
    \vspace{-2.5mm}
    \caption{
    The medians and inter-quartile ranges of the best evaluation values over 20 independent trials.
    }
    \label{fig:MV-BBO_all4d}
\end{figure}

\begin{figure*}[t]
    \centering  \includegraphics[width=0.9\linewidth]{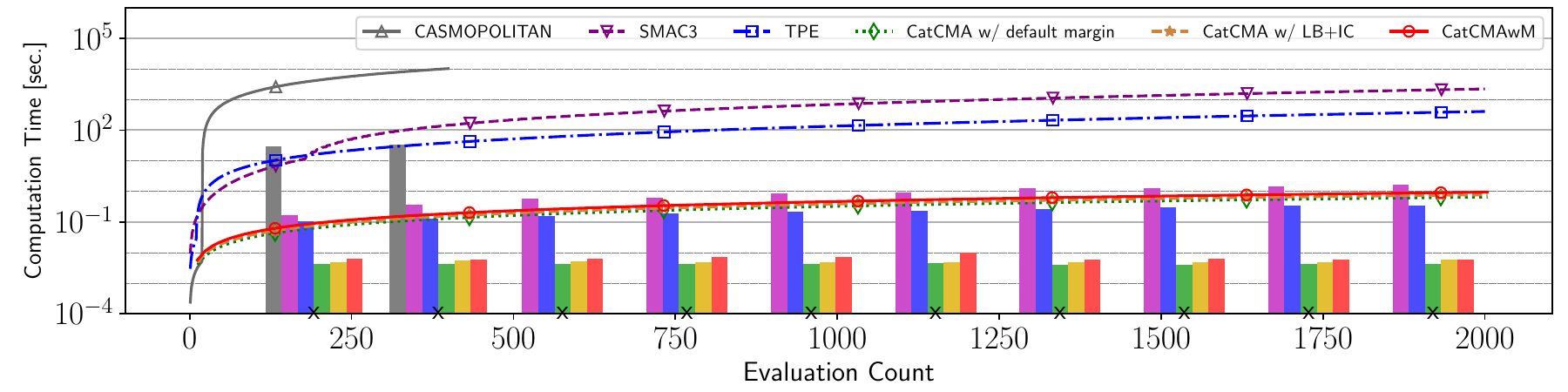}
    \vspace{-1.0mm}
    \caption[time]{Average computation time per iteration (bars) and its cumulative sum (lines) on \textsc{SphereIntCOM} with $(\Nco, \Nin, \Nca) = (6, 6, 6)$ over 20 independent trials. The computational time was measured using an Intel Xeon E5 (3.5 GHz, 6 cores). Optuna version 4.2.0 was used.}
    \label{fig:times}
\end{figure*}

\begin{figure}[t]
    \centering
    \includegraphics[width=0.9\linewidth]{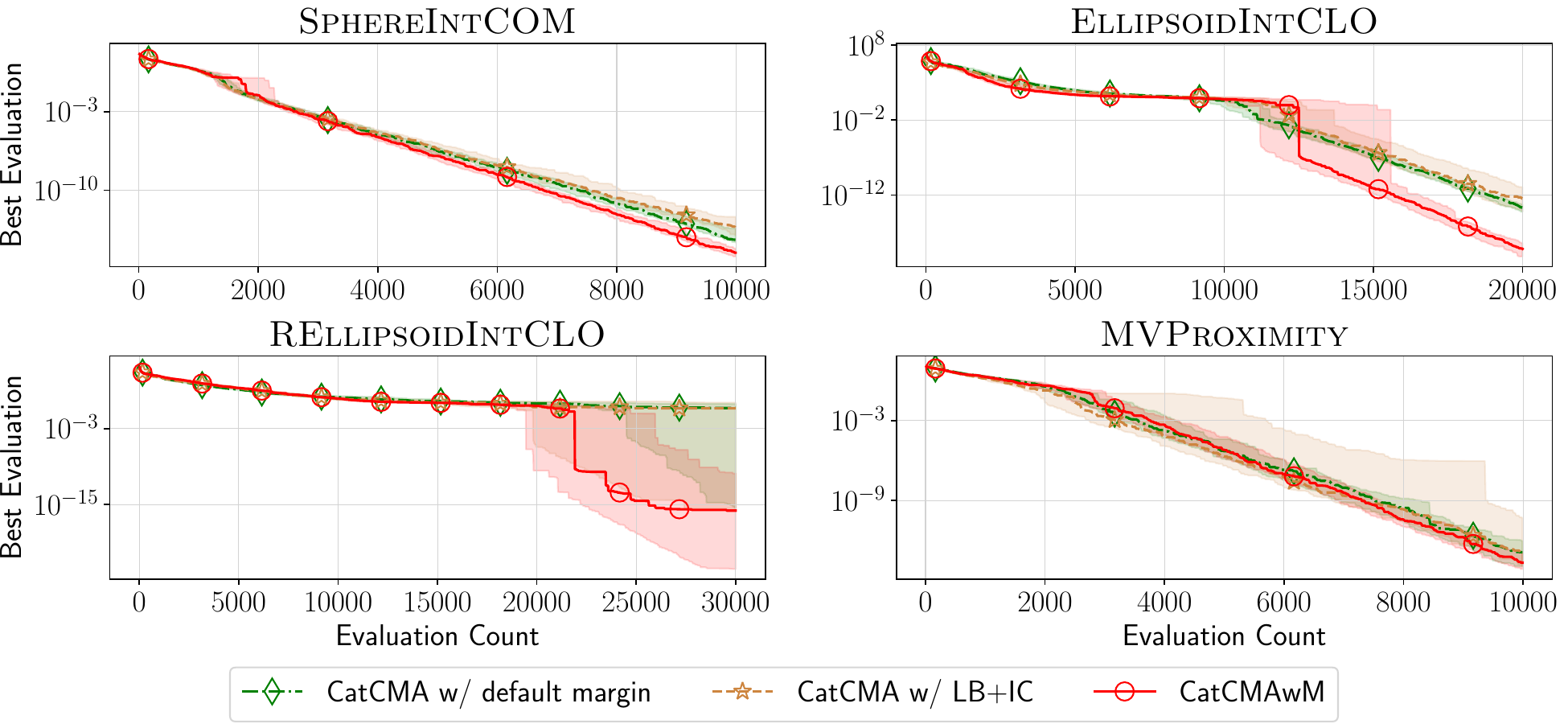}
    \vspace{-2.5mm}
    \caption{The medians and inter-quartile ranges of the best evaluation values for $(\Nco, \Nin, \Nca)=(15,15,15)$ over 20 independent trials.}
    \label{fig:MV-BBO_all4d2}
\end{figure}

\paragraph{Results}
Figure~\ref{fig:MV-BBO_all4d} shows the best evaluation values for each setting.
Except for some cases with low budgets in some settings, CatCMAwM finds better solutions compared to Bayesian optimization methods.
Moreover, as demonstrated in Figure~\ref{fig:times}, CatCMAwM has shorter computation times per iteration and their cumulation compared to Bayesian optimization methods.
Figure~\ref{fig:MV-BBO_all4d2} shows that CatCMAwM achieves superior convergence performance and effectively handles the fixation of discrete variables compared to other CatCMA-based methods.
This is because the modified margin correction sets the maximum margin value such that non-optimal discrete variables do not interfere with the optimization of continuous variables.
Moreover, the results of \textsc{MVProximity} support that CatCMAwM can optimize problems where there are some dependencies among different types of variables.

\section{Extension to Multi-Objective Optimization}
In this section, we extend CatCMAwM from single-objective MV-BBO to the multi-objective setting. We first formalize the multi-objective mixed-variable black-box optimization (MO-MV-BBO) problem and the hypervolume-based performance measure, and then present a Sofomore-based multi-objective extension of CatCMAwM, COMO-CatCMA with Margin.

\subsection{Multi-Objective Mixed-Variable Black-Box Optimization} \label{ssec:MO-MV-BBO}
In this subsection, we introduce the problem formulation and notation for MO-MV-BBO.
The $M$-objective MO-MV-BBO problem is defined as
\begin{align}
    \operatorname*{minimize}_{\x \in \mathcal{X},
    ~\z \in \mathcal{Z}, ~\bc \in {\mathcal{C}}} \bfunc(\x, \z, \bc) = (\bfunc_1(\x, \z, \bc), \ldots, \bfunc_M(\x, \z, \bc)) \enspace,
\end{align}
where $\mathcal{X}$, $\mathcal{Z}$ and $\mathcal{C}$ are defined in Section~\ref{ssec:MV-BBO}.
The goal of multi-objective optimization is to find a diverse set of Pareto-optimal solutions.
In this study, the set of solutions $S$ is evaluated by the hypervolume measure defined as
\begin{align}
    \HV_{\br}(S) = \mathcal{L}_M \left( \bigcup_{(\x, \z, \bc) \in S} \bigl[ \bfunc_1(\x, \z, \bc), \br_1 \bigr] \times \cdots \times \bigl[ \bfunc_M(\x, \z, \bc), \br_M \bigr] \right) \enspace,
\end{align}
where $\br \in \R^M$ is a reference point and $\mathcal{L}_M$ is the Lebesgue measure on the objective space $\R^M$.
In the following, for arbitrary two points $\s, \s'$ in objective space, we denote $\s \prec \s'$ when $\s$ Pareto-dominates $\s'$ and $\s \nprec \s'$ otherwise.
For an arbitrary point $\s$ and a set $S$ in objective space, we also write $S \prec \s$ when there exists $s' \in S$ that dominates $\s$ and $S \nprec \s$ otherwise.

\subsection{Sofomore Framework}
In multi-objective extensions of CMA-ES, a common strategy is to run multiple instances of an elitist CMA-ES variant in parallel \cite{Igel-MO-CMA-ES-1:2007, Igel-MO-CMA-ES-2:2007, improved-MO-CMA-ES:2010}.
This approach has been shown to be effective not only for continuous domains, but also for mixed-integer domains, where an elitist variant of CMA-ES with Margin has been successfully applied \cite{CMAwM-TELO:2024}.
However, for CatCMAwM, which relies on a joint distribution of multivariate Gaussian and categorical distributions, this type of extension is far less straightforward.
The difficulty lies in defining an elite solution for the categorical variables that is consistent with the natural-gradient-based distribution updates.

In this work, we instead base our approach on the Sofomore framework \cite{COMO-CMA-ES:2019}, a general mechanism for constructing multi-objective optimizers from multiple single-objective algorithms without modifying their internal search operators.
Its CMA-ES instantiation, COMO-CMA-ES \cite{COMO-CMA-ES:2019}, has been empirically shown to exhibit linear convergence on a broad class of bi-objective convex quadratic problems and has been reported to achieve competitive or superior performance compared to elitist multi-objective CMA-ES variants, making Sofomore a natural basis for extending CatCMAwM to the multi-objective setting.

\paragraph{Incumbent Solution}
In the Sofomore framework, multiple single-objective optimizers, called \emph{kernels}, are updated one at a time.
The candidate solutions generated by the current kernel are evaluated using a scalar indicator that is computed by taking into account the best estimate of the optimum maintained by all the other kernels.
Then, the Sofomore framework introduces \emph{incumbent solution}, a mapping from the state of each kernel $\theta$ to the solution $\mathcal{E}(\theta)$ that is regarded as its current best estimate of the optimum.
In COMO-CMA-ES, this mapping is instantiated by taking the mean vector of the underlying CMA-ES search distribution as the incumbent solution.

\paragraph{Uncrowded Hypervolume Improvement}
In evolutionary multi-objective optimization, hypervolume improvement (or hypervolume contribution) is commonly used for the fitness of individual solutions.
Given a reference point $\br \in \R^M$, a finite set of solutions $S$, and a candidate solution $\s$, the hypervolume improvement of $\s$ with respect to $S$ is defined as
\begin{align}
    \HVI_{\br} (\s, S) = \HV_{\br} (S \cup \{ \s \}) - \HV_{\br} (S) \enspace.
\end{align}
However, when $\s$ is dominated by $S$, the hypervolume improvement becomes zero.
A common remedy is to combine non-dominated sorting with hypervolume contribution as a secondary criterion, as in SMS-EMOA \cite{SMS-EMOA:2007} and MO-CMA-ES \cite{Igel-MO-CMA-ES-1:2007, Igel-MO-CMA-ES-2:2007, improved-MO-CMA-ES:2010}, but this still tends to attract dominated solutions toward existing non-dominated points rather than into the uncrowded gaps between them.

To solve this problem, the Sofomore framework introduces \emph{uncrowded hypervolume improvement}, which augments hypervolume improvement with a distance-based term that guides dominated solutions toward sparsely populated regions of the empirical Pareto front.
To formalize the uncrowded hypervolume improvement, we first define the empirical Pareto front $\EPF_{S, \br}$ of a finite set $S$ with respect to the reference point $\mathbf{r}$ as
\begin{align}
    \EPF_{S, \br} = \partial U_{S, \br} , \quad \text{with} \enspace U_{S, \br} = \left\{ \boldsymbol{\varsigma} \prec \br \mid \forall \s \in S, \bfunc(\s) \nprec \boldsymbol{\varsigma} \right\} \enspace,
\end{align}
where $\partial U_{S, \br}$ is the boundary of the non-dominated region $U_{S, \br}$.
In addition, the distance to the empirical Pareto front is defined as
\begin{align}
    d_{\br} (\s, S) = \inf_{\boldsymbol{\varsigma} \in \EPF_{S, \br}} d(\bfunc(\s), \boldsymbol{\varsigma}) \enspace,
\end{align}
where $d(\cdot, \cdot)$ denotes the Euclidean distance in objective space.
Finally, combining hypervolume improvement and the distance to the empirical front, the uncrowded hypervolume improvement of $\s$ with respect to $S$ and $r$ is defined as follows.
\begin{align}
    \UHVI_{\br}(\s, S) =
    \begin{cases}
        \HVI_{\br} (\s, S) \quad & \text{if} \enspace \EPF_{S, \br} \nprec \bfunc(\s)  \\
        -d_{\br} (\s, S) \quad & \text{if} \enspace \EPF_{S, \br} \prec \bfunc(\s)
    \end{cases}
\end{align}
Given the current set of incumbents $\Sigma = \{ \s_1, \ldots, \s_p \}$, we define the fitness used to update the $i$-th kernel with respect to other incumbents $\Sigma^{\neg i}$ as
\begin{align}
    \Phi_{\UHVI, \Sigma^{\neg i}} (\s) = \UHVI_{\br}(\s, \Sigma^{\neg i}) \enspace.
\end{align}

\subsection{Proposed Method: COMO-CatCMA with Margin}
In this subsection, we propose COMO-CatCMA with Margin (COMO-CatCMAwM), which instantiates the Sofomore framework with CatCMAwM as the kernel optimizer.
Since CatCMAwM employs the joint probability distribution of multivariate Gaussian and categorical distributions, we need to specify how an incumbent solution is extracted from its internal state.
The mapping function $\mathcal{E}$ is defined as
\begin{align}
    \mathcal{E} : \theta \mapsto \mathcal{E}(\theta) = (\x_{\inc}, \z_{\inc}, \bc_{\inc} ) \enspace,
\end{align}
where $\theta$ denotes the internal state of a CatCMAwM kernel including the mean vector $\boldsymbol{m}$ and the categorical distribution parameter $\q$.
The continuous and integer parts of the incumbent solution are given as follows.
\begin{align}
    [\x_{\inc}]_n &= [\Enc(\boldsymbol{m})]_{\jco_n} \quad \text{for} \enspace n \in \{1, \ldots, \Nco\} \\
    [\z_{\inc}]_n &= [\Enc(\boldsymbol{m})]_{\jin_n} \quad \text{for} \enspace n \in \{1, \ldots, \Nin\}
\end{align}
The categorical part of the incumbent solution is sampled as
\begin{align}
    [\bc_{\inc}]_n \sim \textrm{Uniform} \left( \left\{ \boldsymbol{e}_{n}(k) ~ \middle| ~ k \in \argmax_{k' \in \{ 1, \ldots, K_n \}} \q_{n,k'} \right\} \right) \quad \text{for} \enspace n \in \{1, \ldots, \Nca\} \enspace, \label{eq:incumbent-category}
\end{align}
where $\textrm{Uniform}(\cdot)$ denotes the uniform distribution over a finite set and $\boldsymbol{e}_{n}(k) \in \mathcal{C}_n$ is the one-hot vector corresponding to the $k$-th category of the $n$-th categorical variable.
In most cases, the set in \eqref{eq:incumbent-category} has a single element, and the incumbent categorical value is then selected deterministically.

Algorithm~\ref{alg:COMO-CatCMAwM} shows the optimization process of COMO-CatCMAwM.
When no initial solutions or prior information are provided, each kernel is initialized by sampling its mean vector $\m[0]$ uniformly from the ranges of the continuous and integer variables, and by setting its categorical distribution parameter to the uniform distribution, i.e., for each $n$, $\qt[0]_{n,k} = 1/K_n$ for all $k$.
After initialization, COMO-CatCMAwM proceeds according to the Sofomore framework, as in COMO-CMA-ES. In each iteration, a random permutation of the kernels is generated, and for each kernel in this order a population of candidate solutions is sampled from its current search distribution, evaluated by the UHVI-based scalar objective with respect to the incumbents of the other kernels, and used to perform a CatCMAwM update of the kernel parameters, after which the corresponding incumbent solution and its objective vector are updated.

The code for COMO-CatCMAwM will be made available at \textcolor{blue}{\url{https://github.com/CyberAgentAILab/cmaes}}~\cite{nomura2024cmaes}.

\begin{algorithm}[t!]
    \caption{COMO-CatCMA with Margin}
    \begin{algorithmic}[1]
        \REQUIRE Kernel states $\theta_1, \ldots, \theta_p$, reference point $\br$
        \STATE Initialize incumbents: $\Sigma = \{ \s_1 = \mathcal{E}(\theta_1), \ldots, \s_p = \mathcal{E}(\theta_p) \}$
        \STATE Evaluate $\s_i$ on $\bfunc$ and store $\bfunc(\s_i)$ for $i \in \{1, \ldots, p\}$
        \WHILE{termination conditions are not met}
            \STATE Sample a random permutation $\pi$ of $\{ 1, \ldots, p\}$ uniformly at random
            \FOR{$i=1$ to $p$}
                \STATE Sample $(\x_1, \z_1, \bc_1), \ldots, (\x_\lambda, \z_\lambda, \bc_\lambda)$ from $\pi(i)$-th kernel
                \STATE Evaluate $(\x_1, \z_1, \bc_1), \ldots, (\x_\lambda, \z_\lambda, \bc_\lambda)$ on $\bfunc$
                \STATE Compute the fitness $\Phi(\x_j, \z_j, \bc_j) = \Phi_{\UHVI, \Sigma^{\neg\pi(i)}} (\x_j, \z_j, \bc_j)$ for $j \in \{1, \ldots, \lambda\}$
                \STATE Update $\theta_{\pi(i)}$ according to CatCMAwM using $\Phi(\x_1, \z_1, \bc_1), \ldots, \Phi(\x_\lambda, \z_\lambda, \bc_\lambda)$
                \STATE $\s_{\pi(i)} \leftarrow \mathcal{E}(\theta_{\pi(i)})$
                \STATE Evaluate $\s_{\pi(i)}$ on $\bfunc$ and update the stored objective vector $\bfunc(\s_{\pi(i)})$
            \ENDFOR
        \ENDWHILE
    \end{algorithmic}
    \label{alg:COMO-CatCMAwM}
\end{algorithm}

\subsection{Experiments}
We used DoubleSphereIntLeadingFirstsTrailingLasts (\textsc{DSIntLFTL}) as a bi-objective mixed-variable benchmark function to be minimized.
\begin{align}
    \bfunc_1(\x, \z, \bc) &= \frac{1}{\Nco} \sum_{n=1}^\Nco \left( \frac{\x_n}{x_{\scale}} \right)^2 + \frac{1}{\Nin} \sum_{n=1}^\Nin \left( \frac{\z_n}{z_{\scale}} \right)^2 + \frac{1}{\Nca} \left( \Nca - \sum_{n=1}^{\Nca} \prod_{n'=1}^{n} \bc_{n',1} \right) \\
    \bfunc_2(\x, \z, \bc) &= \underbrace{ \frac{1}{\Nco} \sum_{n=1}^\Nco \left( \frac{\x_n}{x_{\scale}} - 1 \right)^2 + \frac{1}{\Nin} \sum_{n=1}^\Nin \left( \frac{\z_n}{z_{\scale}} - 1 \right)^2 }_{\text{DoubleSphereInt}} + \underbrace{ \frac{1}{\Nca} \left( \Nca - \sum_{n=1}^{\Nca} \prod_{n'=n}^{\Nca} \bc_{n', K_{n'}} \right) }_{\text{LeadingFirstsTrailingLasts}}
\end{align}
\textsc{DSIntLFTL} is the sum of the mixed-integer and categorical benchmark functions. The categorical benchmark, LeadingFirstsTrailingLasts, is designed as an extension of the classical binary benchmark, LeadingOnesTrailingZeros.
Therefore, the Pareto-optimal set of \textsc{DSIntLFTL} consists of solutions whose categorical part has a (possibly empty) prefix of the first category and a (possibly empty) suffix of the last category.

\paragraph{Experimental Settings}
In \textsc{DSIntLFTL}, we set $x_{\smash{\scale}}$ and $z_{\smash{\scale}}$ to $10$.
The ranges of the continuous variables were set to $[-5, 15]$.
The integer space was set as $\mathcal{Z}_1 = \cdots = \mathcal{Z}_{\Nin} = \{-5, -4, \ldots, 14, 15\}$.
The numbers of categories were set to $5$.
The reference point $\br$ was set to $[5, 5]$.
We introduce box-constraint handling~\cite{BoxConst:2009} to CatCMAwM in COMO-CatCMAwM for bounded continuous variables.
As baseline methods, we experimented with NSGA-II~\cite{NSGA-II:2002} and Multiobjective Tree-structured Parzen Estimator (MOTPE)~\cite{MOTPE:2020, MOTPE:2022} implemented by Optuna~\cite{optuna:akiba2019}.
The population size of NSGA-II and the number of initial random solutions in MOTPE were both set to $10$.
In COMO-CatCMAwM, the initial mean vectors of the CatCMAwM kernels were sampled independently and uniformly from $[-5, 15]^{\Nmi}$.
The other distribution parameters were initialized as $\C[0] = \mathbf{I}_{\Nmi}$, $\sig[0] = 4$, and $\boldsymbol{q}_n^{\smash{(0)}} = 1/K_n$.
The kernel size $p$ was set to $10$, and sample size was set as $\lambda = 4 + \lfloor 3 \ln (\Nco + \Nin + \Nca) \rfloor$.
To investigate whether COMO-CatCMAwM is also effective on mixed-integer problems, we additionally consider the setting $K_n = 2$ and compare it against MO-CMA-ES with Margin (MO-CMA-ESwM) \cite{CMAwM-TELO:2024}.
In MO-CMA-ESwM, the parent population size $\lambda$ and the offspring population size $\mu$ were both set to $10$.
The initial mean vectors of MO-CMA-ESwM were set as follows: the components corresponding to continuous and integer variables were sampled independently and uniformly from $[-5, 15]^{\Nmi}$, whereas the components corresponding to categorical (binary) variables were fixed to $(0.5, \ldots, 0.5)$.
The other distribution parameters were initialized as $\C[0] = \mathbf{I}_{\Nmi + \Nca}$ and $\sig[0] = 4$.

\begin{figure*}[t]
    \centering
    \includegraphics[width=0.99\linewidth]{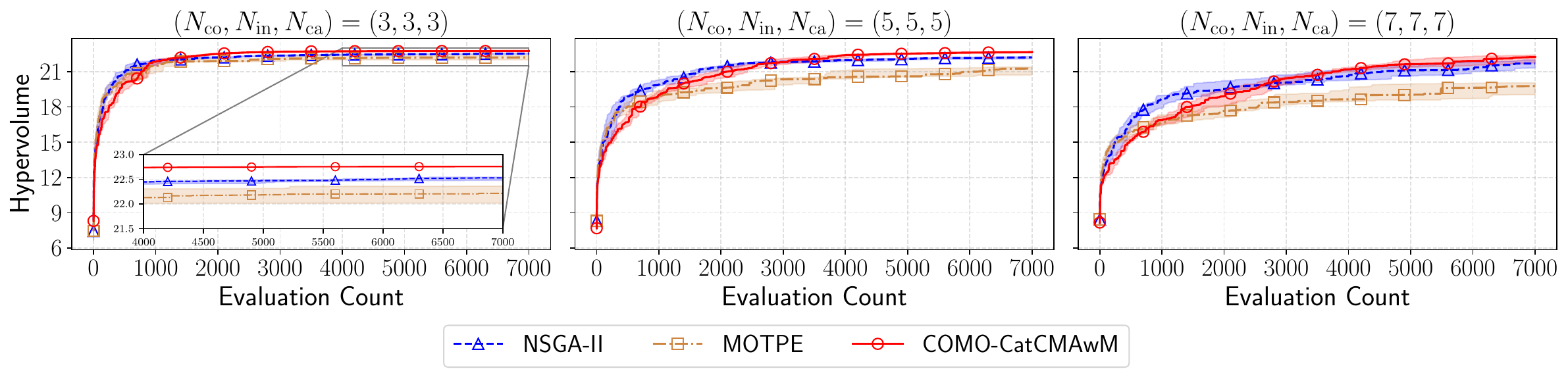}
    \vspace{-1.5mm}
    \caption{The medians and inter-quartile ranges of the hypervolumes for $K_n = 5$ over 20 independent trials.}
    \label{fig:HV}
\end{figure*}

\begin{figure*}[t]
    \centering  \includegraphics[width=0.99\linewidth]{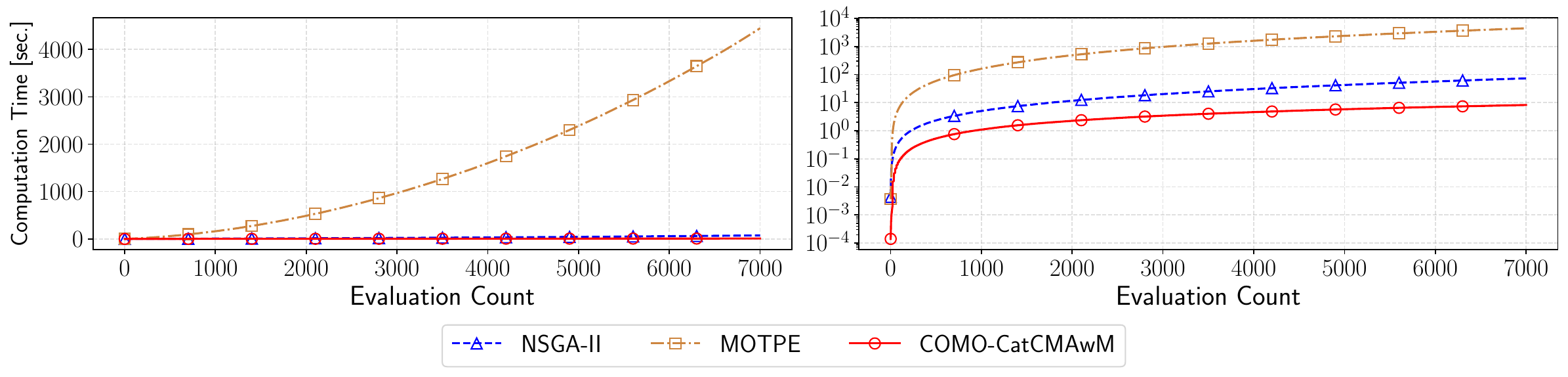}
    \vspace{-1.5mm}
    \caption{Average cumulative computation time for $(\Nco, \Nin, \Nca) = (7,7,7)$, $K_n = 5$ over 20 independent trials. The left and right panels display identical results on a linear and logarithmic scale of the vertical axis, respectively. The computational time was measured using an Intel Xeon E5 (3.5 GHz, 6 cores). Optuna version 4.6.0 was used. The computation time required to compute the hypervolume from the external archive is not included.}
    \label{fig:MO-time}
\end{figure*}

\paragraph{Results}
Figure~\ref{fig:HV} shows the hypervolume values for $K_n = 5$.
The hypervolume is computed from the non-dominated set extracted from an external archive that stores all objective vectors evaluated so far.
Except for small evaluation budgets, COMO-CatCMAwM consistently achieves larger hypervolume values than the baseline methods.
This trend is consistent with the results reported in~\cite{COMO-CMA-ES:2019}, where COMO-CMA-ES generally outperformed NSGA-II in terms of convergence performance; this superior performance was attributed to the UHVI-based fitness and the diversity introduced by the underlying non-elitist evolution strategy.
Figure~\ref{fig:MO-time} shows the average cumulative computation time for $K_n = 5$.
While MOTPE becomes slower as more trials are completed because candidate solutions are generated based on all past evaluations, NSGA-II and COMO-CatCMAwM use the current population or distribution and therefore have lower computational cost.

\begin{figure*}[t]
    \centering
    \includegraphics[width=0.99\linewidth]{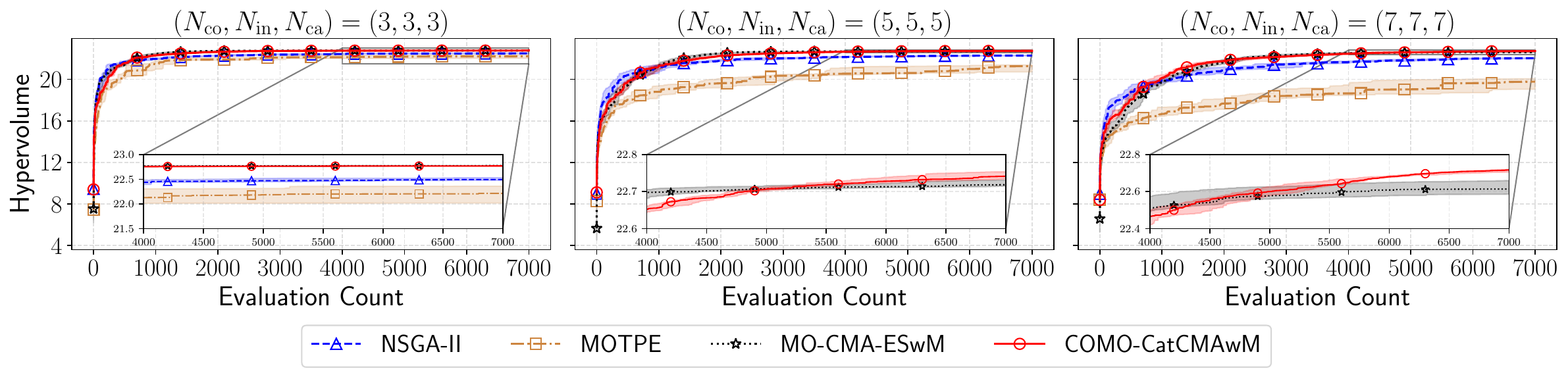}
    \vspace{-1.5mm}
    \caption{The medians and inter-quartile ranges of the hypervolumes for $K_n = 2$ over 20 independent trials.}
    \label{fig:HV-mix}
\end{figure*}

Figure~\ref{fig:HV-mix} shows the hypervolume values for $K_n = 2$, where MO-CMA-ESwM treats categorical variables with $K_n = 2$ as binary variables.
Overall, both MO-CMA-ESwM and COMO-CatCMAwM demonstrate similar performance.
However, COMO-CatCMAwM achieves a slightly higher final hypervolume in the setting of $(N_{\mathrm{co}}, N_{\mathrm{in}}, N_{\mathrm{ca}}) = (7, 7, 7)$.
These results indicate that, even in the mixed-integer case with $K_n = 2$ where the elitist MO-CMA-ESwM is available, the Sofomore-based COMO-CatCMAwM remains competitive.
This suggests that competitive performance may also be attainable without deriving a dedicated elitist extension of CatCMAwM for MO-MV-BBO.
As highlighted in~\cite{COMO-CMA-ES:2019}, a more accurate analysis of convergence performance necessitates evaluating the hypervolume gap relative to a reference value, rather than relying solely on raw hypervolume values.
A more detailed convergence analysis, including the construction of the theoretical Pareto front in mixed-variable search spaces, is left for future work.

\begin{figure*}[t]
    \centering  \includegraphics[width=0.99\linewidth]{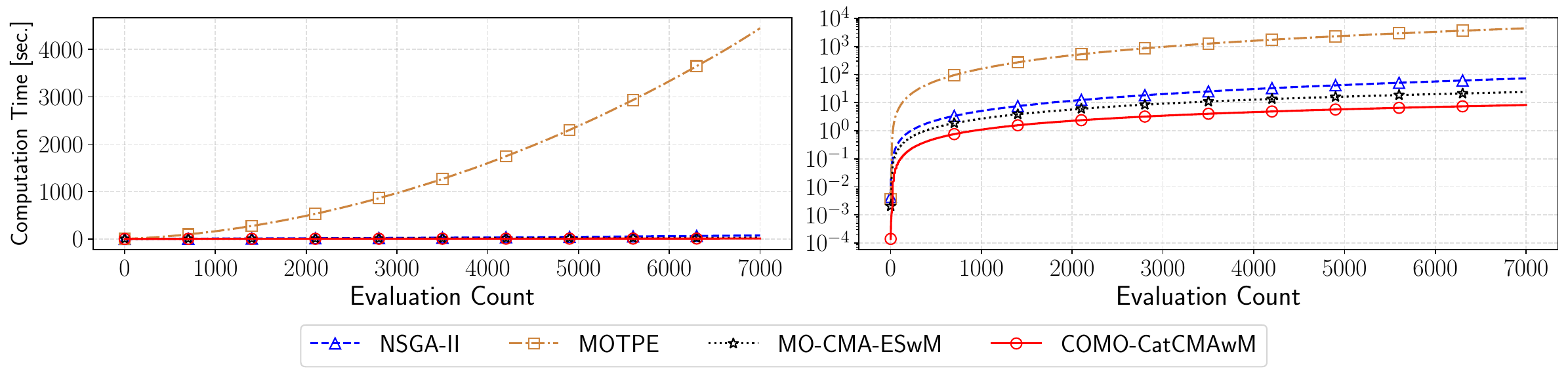}
    \vspace{-1.5mm}
    \caption{Average cumulative computation time for $(\Nco, \Nin, \Nca) = (7,7,7)$, $K_n = 2$ over 20 independent trials. The left and right panels display identical results on a linear and logarithmic scale of the vertical axis, respectively. The computational time was measured using an Intel Xeon E5 (3.5 GHz, 6 cores). Optuna version 4.6.0 was used. The computation time required to compute the hypervolume from the external archive is not included.}
    \label{fig:MO-time-mix}
\end{figure*}

\section{Conclusion}
We proposed CatCMAwM, a stochastic optimization method for continuous, integer, and categorical variables.
CatCMAwM is developed by incorporating a novel integer handling into CatCMA.
The proposed integer handling is carefully designed to work well in CatCMA by focusing on the mutation rate of integer variables.
When applied to mixed-integer problems, it stabilizes the marginal probability and enhances the convergence performance of continuous variables.
Numerical experiments on MV-BBO problems show that the performance of CatCMAwM is superior to the state-of-the-art Bayesian optimization algorithms and CatCMA with naive existing integer handling.

Furthermore, we extended CatCMAwM to the multi-objective setting by instantiating it within the Sofomore framework, obtaining COMO-CatCMAwM for MO-MV-BBO.
Numerical experiments on bi-objective benchmarks show that it achieves competitive or superior hypervolume compared to NSGA-II, MOTPE, and MO-CMA-ESwM.

For future work, we will explore the integration of surrogate models with CatCMAwM, as Bayesian optimization sometimes outperforms our method in low-budget scenarios.
From a benchmarking perspective, benchmark problems that jointly involve continuous, integer, and categorical variables remain scarce in both single- and multi-objective optimization.
Moreover, even for bi-objective mixed-integer problems, the theoretical Pareto fronts are often unavailable, and the underlying problem structure is only partially understood~\cite{tusar_mixed-integer_2019}.
Hence, designing and analyzing various mixed-variable benchmark problems and understanding their theoretical and empirical characterization is an important direction for future work.



\bibliographystyle{ACM-Reference-Format}
\bibliography{reference}

\appendix

\section{Procedure of Modified Margin Correction} \label{apdx:modified-margin-correction}
We consider a successful integer mutation to occur in the dimension $n \in \{1, \ldots, \Nin\}$ when the following condition is satisfied:
\begin{align}
    \exists i \in \{1\!:\!\lambda, \ldots, \mu\!:\!\lambda\} \enspace \text{s.t.} \enspace [\Enc(\x_i)]_{\jin_n} \neq [\Enc(\m[t])]_{\jin_n}
\end{align}
For each $n \in \{1, \ldots, \Nin\}$, the margin correction updates $[\m[t+1]{\mskip -2mu}]_{\jin_n}$ and $\langle \A[t] \rangle_{\jin_n}$ according to whether $n$ is the dimension in which the successful integer mutation occurred.

\paragraph{Case of $~[\textnormal{\textsc{Enc}}(\m[t+1])]_{\jin_n} \in \{ z_{n,1}, z_{n,L_n} \}$.}
Let $\lnei{t+1}{n}$ be the nearest threshold to $\jin_n$-th element of the mean vector. First, the marginal probability $\pmut{t+1}{n}$ is calculated as follows:
\begin{align}
    \pmut{t+1}{n} = &\min\left\{ \Pr\left( [\bv]_{\jin_n} \leq \lnei{t+1}{n} \right),  \Pr\left( \lnei{t+1}{n} < [\bv]_{\jin_n} \right) \right\}
\end{align}
If $n$ is the dimension in which the successful integer mutation occurred, we restrict $\pmut{t+1}{n}$ as
\begin{align}
    \pmut{t+1}{n} \leftarrow \max \left\{ \alpha, \pmut{t+1}{n} \right\} \enspace.
\end{align}
Otherwise, we restrict $\pmut{t+1}{n}$ as
\begin{align}
    \pmut{t+1}{n} \leftarrow \max\left\{ \alpha, \min\left\{ \pmut{t+1}{n}, \pmut{t}{n} \right\} \right\} \enspace.
\end{align}
We correct $\langle \A[t] \rangle_{\jin_n}$ as follows:
\begin{align}
    \langle \A[t+1] \rangle_{\jin_n}  \leftarrow  \max  \left\{  \frac{\left| [\Enc(\m[t+1])]_{\jin_n} - \lnei{t+1}{n} \right|}{\sig[t+1]  \sqrt{\langle \C[t+1] \rangle_{\jin_n}\Xppf{1-2\alpha}}},   \langle \A[t] \rangle_{\jin_n}  \right\}
\end{align}
Finally, we correct $[\m[t+1]]_{\jin_n}$ as follows:
\begin{align}
    \begin{split}
        &[\m[t+1]]_{\jin_n} \leftarrow \rev{\lnei{t+1}{n}} + \sign \left( [\m[t+1]]_{\jin_n} - \rev{\lnei{t+1}{n}} \right) \\
        &\qquad\qquad  \cdot \sig[t+1] \langle \A[t+1] \rangle_{\jin_n} \sqrt{\langle \C[t+1] \rangle_{\jin_n}\Xppf{1-2\pmut{t+1}{n}}}
    \end{split}
\end{align}

\paragraph{Case of $~[\textnormal{\textsc{Enc}}(\m[t+1])]_{\jin_n} \in \{ z_{n,2}, \ldots, z_{n,L_n-1} \}$.}
Let us denote the nearest two thresholds to $[\m[t+1]]_{\jin_n}$ as follows:
\begin{align*}
    \llow{t+1}{n} &= \max \left\{ \ell_{n, l|l+1} \mid l \in \{1, \ldots, L_n-1\}, \ell_{n, l|l+1} < [\m[t+1]]_{\jin_n} \right\} \\
    \lup{t+1}{n} &= \min \left\{ \ell_{n, l|l+1} \mid l \in \{1, \ldots, L_n-1\}, [\m[t+1]]_{\jin_n} \leq \ell_{n, l|l+1} \right\}
\end{align*}
The marginal probabilities are calculated as
\begin{align}
    \plow{t+1}{n} &= \Pr\left( [\bv]_{\jin_n} \leq \llow{t+1}{n} \right) \enspace, \\
    \pup{t+1}{n} &= \Pr\left( \lup{t+1}{n} < [\bv]_{\jin_n} \right) \enspace, \\
    \pmid{t+1}{n} &= 1 - \plow{t+1}{n} - \pup{t+1}{n} \enspace.
\end{align}
If $n$ is the dimension in which the successful integer mutation occurred, the marginal probabilities are modified as follows:
\begin{align}
    \plow{t+1}{n} &\leftarrow \max\left\{\frac{\alpha}{2}, \plow{t+1}{n}\right\} \\
    \pup{t+1}{n} &\leftarrow \max\left\{\frac{\alpha}{2}, \pup{t+1}{n}\right\} \\
    \Delt{t+1}{n} &\leftarrow \frac{1 - \plow{t+1}{n} - \pup{t+1}{n} - \pmid{t+1}{n}}{\plow{t+1}{n} + \pup{t+1}{n} + \pmid{t+1}{n} - 3\cdot \alpha/2} \\
    \plow{t+1}{n} &\leftarrow \plow{t+1}{n} + \Delt{t+1}{n} (\plow{t+1}{n} - \alpha/2) \\
    \pup{t+1}{n} &\leftarrow \pup{t+1}{n} + \Delt{t+1}{n} (\pup{t+1}{n} - \alpha/2)
\end{align}
Otherwise, the marginal probabilities are modified as follows:
\begin{align}
    \plow{t+1}{n} &\leftarrow \max\left\{\frac{\alpha}{2}, \plow{t+1}{n}\right\} \\
    \pup{t+1}{n} &\leftarrow \max\left\{\frac{\alpha}{2}, \pup{t+1}{n}\right\}  \\
    \pmid{t+1}{n} &\leftarrow \max\left\{1 - \pmut{t}{n}, \pmid{t+1}{n}\right\}  \\
    \Delt{t+1}{n} &\leftarrow \frac{1 - \plow{t+1}{n} - \pup{t+1}{n} - \pmid{t+1}{n}}{\plow{t+1}{n} + \pup{t+1}{n} + \pmid{t+1}{n} - \alpha - (1 - \pmut{t}{n})} \\
    \plow{t+1}{n} &\leftarrow \plow{t+1}{n} + \Delt{t+1}{n} (\plow{t+1}{n} - \alpha/2) \\
    \pup{t+1}{n} &\leftarrow \pup{t+1}{n} + \Delt{t+1}{n} (\pup{t+1}{n} - \alpha/2)
\end{align}
We calculate $\pmut{t+1}{n}$ as
\begin{align}
    \pmut{t+1}{n} \leftarrow \plow{t+1}{n} + \pup{t+1}{n} \enspace.
\end{align}
Finally, $[\m[t+1]]_{\jin_n}$ and $\langle\A[t+1]\rangle_{\jin_n}$ are corrected as follows:
\begin{align*}
    [\m[t+1]]_{\jin_n} &\leftarrow \frac{\llow{t+1}{n} \sqrt{\Xppf{1-2\pup{t+1}{n}}} + \lup{t+1}{n} \sqrt{\Xppf{1-2\plow{t+1}{n}}}}{\sqrt{\Xppf{1-2\pup{t+1}{n}}} + \sqrt{\Xppf{1-2\plow{t+1}{n}}}} \\
    \langle\A[t+1]\rangle_{\jin_n} &\leftarrow \frac{\lup{t+1}{n} - \llow{t+1}{n}}{\sig[t+1]\sqrt{\langle\C[t+1]\rangle_{\jin_n}}\left(\sqrt{\Xppf{1-2\pup{t+1}{n}}} + \sqrt{\Xppf{1-2\plow{t+1}{n}}}\right)}
\end{align*}

\section{Proof of Proposition~\ref{prop:int_margin}} \label{apdx:proof-int-margin}
\begin{proof}
    When the distribution parameters satisfy \eqref{eq:mix_prop_cond1} and \eqref{eq:mix_prop_cond2}, the probability that at least one non-optimal integer is included in a sample is calculated as follows:
    \begin{align}
        1 - \prod_{n=1}^{\Nin} (1-\pmut{t}{n}) & = 1 - \prod_{n=1}^{\Nin} \left(1-(1 - 0.73^\frac{1}{\Nin}) \right) \\
        & = 1 - \prod_{n=1}^{\Nin} 0.73^\frac{1}{\Nin} \\
        & = 0.27
    \end{align}
    Then, the random variable $\lambda_\non$ follows the binomial distribution $\textrm{Bin}(\lambda, 0.27)$.
    From the lower bound of $\Pr (\lambda_\non \leq \lambda - \lfloor \lambda / 2 \rfloor)$ shown in the proof of \cite[Proposition~4.1]{CatCMA:2024}, we have proven $\Pr (\lambda_\non \leq \lambda - \lfloor \lambda / 2 \rfloor) \geq 0.95$ when $\lambda \geq 6$.
\end{proof}

\section{Proof of Proposition~\ref{prop:mix_margin}} \label{apdx:proof-mix-margin}
\begin{proof}
    When the distribution parameters satisfy \eqref{eq:int_prop_cond1}, \eqref{eq:int_prop_cond2}, and \eqref{eq:assum_cat}, the probability that at least one non-optimal integer and/or one non-optimal category is included in a sample is calculated as
    \begin{align}
        1 - \prod_{n=1}^{\Nin} (1-\pmut{t}{n}) \cdot \prod_{n=1}^{\Nca} \qt[t]_{n,1} \enspace, \label{eq:non_optimal_prob}
    \end{align}
    where $\prod_{n=1}^{\Nin} (1-\pmut{t}{n})$ is calculated as follows:
    \begin{align}
        \prod_{n=1}^{\Nin} (1-\pmut{t}{n}) & = \prod_{n=1}^{\Nin} \left(1-(1 - 0.73^\frac{1}{\Nin+\Nca}) \right) \\
        & = \prod_{n=1}^{\Nin} 0.73^\frac{1}{\Nin+\Nca}
    \end{align}
    and $\prod_{n=1}^{\Nca} \qt[t]_{n,1}$ is calculated as follows:
    \begin{align}
        \prod_{n=1}^{\Nca} \qt[t]_{n,1} &= \prod_{n=1}^{\Nca} \left(1-\frac{1}{K_n - 1}\left( 1 - 0.73^\frac{1}{\Nin+\Nca} \right) (K_n - 1) \right) \\
        &= \prod_{n=1}^{\Nca} 0.73^\frac{1}{\Nin+\Nca}
    \end{align}
    Then, \eqref{eq:non_optimal_prob} can be calculated as follows:
    \begin{align}
        &1 - \prod_{n=1}^{\Nin} (1-\pmut{t}{n}) \cdot \prod_{n=1}^{\Nca} \qt[t]_{n,1} \notag \\
        &\qquad = 1 - \prod_{n=1}^{\Nin} 0.73^\frac{1}{\Nin+\Nca} \cdot \prod_{n=1}^{\Nca} 0.73^\frac{1}{\Nin+\Nca} \\
        &\qquad = 0.27
    \end{align}
    Then, the random variable $\lambda_\non$ follows the binomial distribution $\textrm{Bin}(\lambda, 0.27)$.
    From the lower bound of $\Pr (\lambda_\non \leq \lambda - \lfloor \lambda / 2 \rfloor)$ shown in the proof of \cite[Proposition~4.1]{CatCMA:2024}, we have proven $\Pr (\lambda_\non \leq \lambda - \lfloor \lambda / 2 \rfloor) \geq 0.95$ when $\lambda \geq 6$.
\end{proof}

We note that there are other values of $\alpha$ and $\qmin_n$  such that equation \eqref{eq:non_optimal_prob} equals to $0.27$, such as $\alpha = 1 - 0.73^{\frac{1}{2\Nin}}$ and $\qmin_n = \frac{1}{K_n - 1} \left( 1 - 0.73^{\frac{1}{2\Nca}} \right)$.
Considering the case of $\Nin = 0$ or $\Nca = 0$, we set $\alpha = 1 - 0.73^{\frac{1}{\Nin + \Nca}}$ and $\qmin_n = \frac{1}{K_n - 1} \left( 1 - 0.73^{\frac{1}{\Nin + \Nca}} \right)$.
The discussion of this optionality is future work.

\end{document}